\documentclass{article}

\usepackage{listings}
\usepackage{xcolor}
\usepackage{float}
\usepackage{graphicx}
\usepackage{subcaption}
\usepackage{multirow}
\usepackage{makecell}
\usepackage{threeparttable}
\usepackage{arxiv}
\usepackage{amsmath}
\usepackage{array}
\usepackage[utf8]{inputenc} 
\usepackage[T1]{fontenc}    
\usepackage{newunicodechar}
\newunicodechar{ }{~} 
\usepackage{hyperref}       
\usepackage{url}            
\usepackage{booktabs}       
\usepackage{amsfonts}       
\usepackage{nicefrac}       
\usepackage{microtype}      
\usepackage{lipsum}		
\usepackage{graphicx}
\usepackage{natbib}
\usepackage{doi}
 \usepackage{placeins}

\title{Strategic Intelligence in Large Language Models\\
Evidence from evolutionary Game Theory.}

\author{ \href{https://orcid.org/0000-0002-4688-9285}{\includegraphics[scale=0.06]{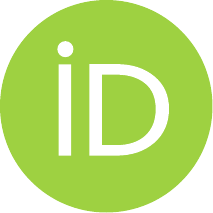}\hspace{1mm}Kenneth Payne} \\
	King's College London\\
	\texttt{kenneth.payne@kcl.ac.uk}\\
	\And
	\href{https://orcid.org/0009-0008-8108-3317}{\includegraphics[scale=0.06]{orcid.pdf}\hspace{1mm}Baptiste Alloui-Cros} \\
	University of Oxford \\
	\texttt{baptiste.alloui-cros@politics.ox.ac.uk} \\
}

\renewcommand{\headeright}{Preprint}
\renewcommand{\undertitle}{}
\renewcommand{\shorttitle}{}

\hypersetup{
pdftitle={Strategic Intelligence in Large Language Models: Evidence from Evolutionary Game Theory.},
pdfsubject={q-bio.NC, q-bio.QM},
pdfauthor={Kenneth Payne, Baptiste Alloui-Cros},
pdfkeywords={First keyword, Second keyword, More},
}

\begin{document}
\maketitle
\begin{abstract}

Are Large Language Models (LLMs) a new form of strategic intelligence, able to reason about goals in competitive settings? We present compelling supporting evidence. The Iterated Prisoner's Dilemma (IPD) has long served as a model for studying decision-making. We conduct the first ever series of evolutionary IPD tournaments, pitting canonical strategies (e.g., Tit-for-Tat, Grim Trigger) against agents from the leading frontier AI companies OpenAI, Google, and Anthropic. By varying the termination probability in each tournament (the "shadow of the future"), we introduce complexity and chance, confounding memorisation.

Our results show that LLMs are highly competitive, consistently surviving and sometimes even proliferating in these complex ecosystems. Furthermore, they exhibit distinctive and persistent "strategic fingerprints": Google's Gemini models proved strategically ruthless, exploiting cooperative opponents and retaliating against defectors, while OpenAI's models remained highly cooperative, a trait that proved catastrophic in hostile environments. Anthropic's Claude emerged as the most forgiving reciprocator, showing remarkable willingness to restore cooperation even after being exploited or successfully defecting. Analysis of nearly 32,000 prose rationales provided by the models reveals that they actively reason about both the time horizon and their opponent's likely strategy, and we demonstrate that this reasoning is instrumental to their decisions. This work connects classic game theory with machine psychology, offering a rich and granular view of algorithmic decision-making under uncertainty.

\end{abstract}


\section{Introduction}

We explore whether modern AI agents exhibit strategic decision-making amidst uncertainty. Iterated Prisoner's Dilemma (IPD) tournaments have long been a work‑horse for the quantitative study of cooperation and strategy evolution. The Prisoner's Dilemma is a widely explored game theory scenario where the individually rational move (defect) yields sub-optimal outcomes for both participants. Playing the game repeatedly, however, makes reputation and retaliation important parts of the decision-making calculation—defection might now be punished in future rounds. Axelrod's seminal computer tournaments in the late 1970s and early 1980s showed how a simple reciprocal rule—Tit‑for‑Tat—could out‑compete far more elaborate heuristics in repeated encounters, spawning an extensive literature on evolutionary game theory, bounded rationality, and the conditions under which cooperation flourishes.

Subsequent experiments have extended Axelrod's paradigm along multiple dimensions: introducing noise (Nowak \& Sigmund, 1990), varying population structures and network connectivity (Ifti et al., 2004), allowing for varied game horizons (Dal Bó, 2005), or incorporating more sophisticated learning algorithms such as Q‑learning and genetic programming (Sandholm \& Crites, 1996).Yet two features persist: (i) the agents are either hand‑coded or governed by relatively low‑capacity algorithms, and (ii) the agents’ internal reasoning—how they \textit{explain} their choices—remains either opaque or, in the case of deterministic agents, entirely absent.

Meanwhile, with large language models (LLMs), a new type of reasoning agent has arrived, with potentially revolutionary implications for decision-making amidst uncertainty. The nascent field of ‘machine psychology’ (Hagendorff et al., 2023) sets out to explore the extent to which these models display human-like decision-making characteristics, or otherwise (Strachan et al., 2024). Scholars are particularly interested in whether the models are genuinely capable of reasoning. This creates an opportunity to revisit the IPD, deploying with LLM agents, to explore how far they are capable of natural‑language reasoning, on‑the‑fly adaptation, and meta‑cognitive reflection. LLM‑driven agents can process a formal system prompt (e.g., the payoff matrix and termination probability), generate a textual justification for each move, and potentially revise their play in light of that reflection. This brings the experimental set‑up closer to the human behavioural experiments that followed Axelrod’s tournaments, while retaining the scale and controllability of simulation.

\subsection{Contribution}

We report the first evolutionary IPD tournament populated by classic benchmark strategies \textit{and} contemporary LLM agents (using models from OpenAI, Google and Anthropic). We first implement a 2 × 2 combination of experiments that crosses model capability (basic vs advanced) with the "shadow of the future" (10 \% vs 25 \% per‑round termination probability). We augment these results with two additional tournaments, both on more advanced models, and designed as stress tests: a 75 \% termination regime (collapsing the iterated aspect to near, but not quite, one‑shot); and a persistent mutation regime that re‑injects a Random agent each phase to avoid settled equilibria – bringing the tournament closer to biological evolution, albeit still heavily stylised. Lastly, we conduct an AI-heavy tournament involving three LLMs from rival companies against a Bayesian algorithm that performed effectively throughout. In all we conduct seven round-robin tournaments, producing almost 32,000 individual decisions and rationales from the language models.\

Our aims are threefold:
\begin{enumerate}
    \item Benchmark LLM strategies against canonical baselines. Can language models compete effectively in the competitive and uncertain world of IPD? We find strong evidence that they are effective in a broad range of conditions, with more advanced models performing better than comparatively basic ones.
    
    \item Explore the ways in which the models approach the tournaments. Are advanced LLMs merely memorisers, or ‘stochastic parrots’, deterministically predicting their output on the basis of training data? Do they model tit‑for‑tat because it played well in the scholarly literature? Or do they display strategic \textit{fingerprints}, with their own distinctive playing styles? We find strong evidence for the latter, with a Machiavellian Gemini, in particular, proving more adaptable than the broadly trusting OpenAI.
 
\item Probe the impact of horizon length and adversary modelling on evolutionary success. Does the classic shadow‑of‑the‑future logic feature when agents articulate their reasoning? We find strong evidence that models perform differentially under different shadow-lengths, and that they explicitly factor it into their rationales. Similarly, we find abundant evidence that the models consider and respond to adversary behaviour when making their decisions – in many instances, figuring out what classic agent they are facing, and adjusting accordingly. Again, this is particularly true of Gemini.

\end{enumerate}

By coupling population‑level evolutionary metrics with natural‑language justifications, we offer a richer picture of algorithmic cooperation than has been possible with traditional hand‑coded agents. Our experiments and our results speak both to the game‑theoretic community, since we test long‑standing hypotheses under a new class of strategies, and to the emerging field of ‘machine psychology’, which explores the ways in which modern AI agents, especially transformers, interact with each other, and humans. 

 A key debate in the literature on LLMs is the extent to which they reason, versus merely retrieving memorized patterns. Some foundational work suggests that LLMs possess latent reasoning abilities that can be elicited with specific prompting techniques (Wei et al., 2022), and have even been shown to spontaneously develop sophisticated cognitive capabilities such as a Theory of Mind (Kosinski, 2023). In contrast, others provide a  skeptical viewpoint, arguing that LLM performance can be explained by the memorization of training datasets, rather than reasoning (Razeghi et al., 2022; Kandpal et al., 2023).  Some work complicates this binary, suggesting that memorization may be a necessary intermediate step on the path to true generalization (Power et al., 2022), a phenomenon termed 'grokking'.
 
 We contribute to this debate by deploying agents in situations where memorisation, including of the scholarly literature on IPD is of limited utility in informing their decisions. That's especially true when playing against other LLM agents - a strategic interaction that does not yet exist in the literature. We also confound the utility of memorisation by introducing noise into the tournaments, via a Random agent, and in one run via mutation. And of course we add further challenges for memorisation by introducing and varying uncertainty about when each match will terminate – a so-called ‘shadow-of-the future’ which alters the incentives for cooperation. All these factors make it extremely challenging to definitively determine the winning strategy \textit{ex ante}, solely via retrieval from training data. Since they cannot readily draw on memories of the literature to guide their on-the-fly decisions, we conclude that this reasoning is integral to their decision, rather than a spandrel (in evolutionary psychology, an evolved cognitive artefact that serves no instrumental purpose). Specifically, we find that the models often reason along two axes – first, about the time horizon of the tournament, which alters the incentives to cooperate or defect; and, secondly, about the likely strategy of their adversaries, by examining their prior moves and conjecturing about their future moves on that basis. Then they decide.

 Our major finding is that LLMs are competitive in all variations of the tournament. They demonstrate considerable ability, such that they are almost never eliminated by the fitness selection criteria (the exception is one tournament where Gemini overwhelms all but one of its rivals, including the OpenAI agent). We also find that LLMs exhibit different strategic styles, or fingerprints, which have an impact on their success in varying conditions. OpenAI's models are consistently more cooperative than Gemini, which proves able to adapt successfully to changing environmental conditions. Later, we see that Anthropic's Claude is more cooperative still, but nonetheless outperforms OpenAI head-to-head 

\section{Method}

\subsection{Experimental Design}
We implemented a series of evolutionary tournaments, each consisting of five phases in which a population of agents engages in round-robin Iterated Prisoner’s Dilemma (IPD) matches. After each phase, agents reproduce in proportion to their average per move score, creating a new population for the next phase. The design is summarised in Table~\ref{tab:experimental_factors}.

\begin{table}[h!]
\centering
\caption{Overview of Experimental Factors and Levels}
\label{tab:experimental_factors}
\begin{tabular}{>{\raggedright\arraybackslash}p{0.2\textwidth} >{\raggedright\arraybackslash}p{0.4\textwidth} >{\raggedright\arraybackslash}p{0.35\textwidth}}
\toprule
\textbf{Factor} & \textbf{Levels} & \textbf{Purpose} \\
\midrule
Model capability & Basic (\texttt{gpt-3.5-turbo}, \texttt{gemini-1.5-flash-preview-0514}) vs Advanced (\texttt{gpt-4o-mini}, \texttt{gemini-2.5-flash}) & Tests whether larger, more recent LLMs outperform earlier versions in both pay offs and evolutionary survival. \\
\addlinespace

Shadow of the future & 10\% vs 25\% per round termination probability & Varies expected match length; classical theory predicts cooperation should fall as termination probability rises. \\
\addlinespace

Stress tests & 75\% termination \& Random mutation regime & Pushes agents to near one shot play and tests population stability under continual perturbation. \\
\addlinespace

LLM showdown & 10\% termination; 3 relatively advanced LLMs (\texttt{Claude-3-Haiku}, \texttt{gemini-2.5-flash}, \texttt{gpt-4o-mini}) & Concentrates the AI interactions, and increases the novelty of the encounters. \\
\bottomrule
\end{tabular}
\end{table}

The core experiment is therefore a 2 × 2 factorial design (model capability × termination probability). Each cell is executed once, yielding 276 round-robin matches per phase (with a fixed population of 24 agents) and a total of five phases per tournament. At the end of each phase, the best performing agents may increase in frequency for the next phase, at the expense of the worst, allowing us to explore evolutionary dynamics of cooperation and reputation. In the final, seventh, IPD tournament we pitted three LLMs against one another, introducing a comparably sophisticated/recent model from rival company, Anthropic.

\subsection{Match Procedure}

\begin{enumerate}

\item The rewards in each stage are the classic PD matrix R = 3, S = 0, T = 5, P = 1, where R is the reward; S is sucker; T is temptation and P is punishment.

\begin{table}[h!]
\centering
\caption{The Prisoner's Dilemma Payoff Matrix. Payoffs are listed as (Player 1, Player 2).}
\label{tab:payoff_matrix}
\begin{tabular}{lcc}
\toprule
& \textbf{Player 2 Cooperates} & \textbf{Player 2 Defects} \\
\midrule
\textbf{Player 1 Cooperates} & (3, 3) & (0, 5) \\
\textbf{Player 1 Defects}    & (5, 0) & (1, 1) \\
\bottomrule
\end{tabular}
\end{table}

\item Initial move - Each agent receives an empty history and the termination probability p. In subsequent moves, the agents receive the match history of prior moves, up to a maximum of 20 moves.

\item Iterated rounds -- After each pair of simultaneous moves the match ends with probability \(p\), or once a hard cap of 30 rounds is reached. The probability of reaching this cap is \(\approx 4.7\%\) when \(p = 0.10\), \(\approx 0.024\%\) when \(p = 0.25\), and \(< 10^{-16}\) when \(p = 0.75\).

\end{enumerate}

Each phase therefore contains \(n(n-1)/2\) matches; with \(n = 24\) agents this yields the 276 matches noted above. With \(p \in \{0.10, 0.25, 0.75\}\), the expected match lengths are \(9 \pm 3\), \(3 \pm 1\), and \(1.3 \pm 0.1\) rounds respectively.\\

\subsection{Agent Set}

\subsubsection{Classic strategies}

We include ten canonical IPD strategies, custom coded in python, and implemented as described in Axelrod (1984) and subsequent literature:

\begin{itemize}
 \item \textbf{Tit for Tat} (TFT): This is the most famous and historically successful strategy for the Iterated Prisoner's Dilemma. It begins by cooperating on the first move. Thereafter, it simply copies the opponent's move from the previous round. This strategy is "nice" because it starts by cooperating, "retaliatory" because it immediately punishes defection, and "forgiving" because it will cooperate again as soon as the opponent does.

\item \textbf{Grim Trigger} (GT): This strategy is also known as "Trigger" or "Grim". It starts by cooperating and continues to cooperate as long as the opponent cooperates. However, if the opponent defects even once, Grim Trigger will defect for the remainder of the match without exception. It is nice and retaliatory, but completely unforgiving.

\item \textbf{Win-Stay, Lose-Shift} (WSLS): Also known as Pavlov, this strategy bases its decision on the outcome of the previous round. It defines a "win" as receiving one of the two higher payoffs (Temptation: 5, or Reward: 3) and a "loss" as receiving one of the lower payoffs (Punishment: 1, or Sucker: 0). If it won in the previous round, it repeats its move. If it lost, it switches its move. It starts by cooperating. This allows it to correct for occasional mistakes and exploit unconditionally cooperative opponents.

\item \textbf{Generous Tit for Tat} (GenerousTFT): This is a variant of Tit for Tat designed to be more resilient in noisy environments where accidental defections can occur. Like TFT, it cooperates on the first move and generally reciprocates the opponent's last move. However, if the opponent defects, GenerousTFT has a 10\% probability of "forgiving" the defection and cooperating in the next round. This can break the "death spiral" of mutual defections that can occur between two TFT agents if one accidentally defects.

\item \textbf{Suspicious Tit for Tat} (SuspiciousTFT): This is the inverse of Tit for Tat's "nice" starting move. It begins by defecting on the first move to test the opponent. After the first move, it behaves exactly like Tit for Tat, copying the opponent's previous move. This strategy aims to exploit overly optimistic opponents without getting locked into long-term defection against retaliatory ones.

\item
  \textbf{Prober}: This strategy, sometimes called Tit for Two Tats, acts as a
  probe to identify the opponent's nature. It
  starts with a fixed sequence: Cooperate, Defect, Cooperate. If the
  opponent cooperates in response to Prober's defection on the second move
  and continues cooperating on the third move, Prober concludes the opponent is cooperative and switches to a standard Tit for Tat strategy for the
  rest of the match. If the opponent defects in response to Prober's test
  defection, Prober concludes the opponent is uncooperative and defects for
  the rest of the match.

\item \textbf{Random}: This agent makes its decision randomly, with a 50\% chance of cooperating and a 50\% chance of defecting on each move, regardless of the history of the game. It serves as a baseline control in the simulation.

\item \textbf{Gradual}: This is a more complex retaliatory strategy. It starts by cooperating. If the opponent defects, it responds with a number of defections equal to the total number of times the opponent has defected so far. After delivering these punishments, it cooperates for two rounds to signal a willingness to return to mutual cooperation. It is a forgiving strategy, but its punishment is proportional to the opponent's transgressions.

\item \textbf{Alternator}: This agent follows a simple, deterministic pattern of alternating between cooperation and defection. It starts with Cooperation, then Defects, then Cooperates, and so on (C, D, C, D...).

\item \textbf{Bayesian}: This is the most computationally intensive of the classic strategies. It attempts to infer the opponent's strategy from its moves. It maintains a probability distribution over a set of known, simple strategies (specifically: Tit for Tat, Grim Trigger, Always Cooperate, and Always Defect). After each move, it uses Bayes' theorem to update the probabilities based on the opponent's last action. It then selects its own move by choosing the best response to the strategy that is currently considered most likely.

\end{itemize} 

\subsubsection{LLM agents}
Our tournaments featured two types of LLM-driven agents, each used in two different configurations: a "basic" model and a more "advanced" model.

\begin{itemize}
    \item \textbf{Basic Models}: Google's \texttt{gemini-1.5-flash} and OpenAI's \texttt{gpt-3.5-turbo}.
    \item \textbf{Advanced Models}: Google's \texttt{gemini-2.5-flash} and OpenAI's \texttt{gpt-4o-mini}. In the three-way showdown, we also included Anthropic's \texttt{claude-3-haiku-20240307}.
\end{itemize}

While the models in each tier are functional equivalents and serve similar roles in their respective ecosystems, they are not architecturally identical. This deliberate choice is central to our experiment, as it allows us to investigate whether distinct underlying training data and/or fine-tuning methods result in divergent strategic ‘fingerprints’.

To isolate model effects, prompts for the models were token-identical. For the advanced OpenAI models, the sampling temperature was set to 0.7. For the Gemini models, the temperature was left at its API default. All other sampling parameters were left at their respective defaults. No external memory beyond the current match history was provided. The initial tournament population is set at 2 copies of each agent, for n=24. 

For the last tournament, we introduced a third LLM, Anthropic’s claude-3-haiku-20240307, broadly comparable in terms of overall model spec and performance, to provide an additional comparison for the benchmarks set by the two core models. For consistency, Claude's sampling temperature was set to 0.7 and all other sampling parameters were left at their respective API defaults.

Each LLM agent received a standardized prompt for every move decision, including: (1) game rules and payoff matrix, (2) current termination probability, (3) complete paired move history with the current opponent, and (4) instructions to provide reasoning followed by a single-letter move decision (C or D). The prompt explicitly stated the goal of maximizing total score and provided the standard Prisoner's Dilemma payoffs. We furnished termination probability information to test agent capacity for strategic reasoning with complete information, rather than their ability to infer game parameters. This design choice allows us to isolate strategic reasoning abilities from parameter discovery, focusing on how these models utilize known strategic information.

Here is an illustrative example of the rationales the models produced. Gemini is playing GenerousTitforTat and defecting. This happened in move 4 of match 170 in phase 1 of the tournament with a 25\% shadow of the future.

\begin{itemize}
    \item Gemini: "\textit{The opponent has defected twice and cooperated once. Since there's a 25\% chance the game ends after each round, I should prioritize maximizing my points in the short term. Defecting is the best strategy for maximizing points, especially since the opponent seems likely to defect as well.}"
\end{itemize}

\subsection{Evolutionary Update Rule}

Our reproduction procedure is designed to amplify selection pressure. It is based on a strategy's performance relative to the average performance of all unique strategies present in a given phase.

First, we define the average score per move for a given strategy \(i\) in phase \(t\), which we term its fitness, \(F_{i,t}\). This is calculated by dividing the total score accumulated by all agents of strategy \(i\) in phase \(t\) (\(S_{i,t}\)) by their total number of moves in that phase (\(M_{i,t}\)):

\begin{equation}
F_{i, t} = \frac{S_{i,t}}{M_{i,t}}
\end{equation}

Next, we calculate the mean fitness for the phase, \(\bar{F_t}\). This is the \textbf{unweighted average} of the fitness scores of all unique strategies that were present in the population during phase \(t\) (\(P_t\)). If there are \(k\) unique strategies in \(P_t\), the mean fitness is:

\begin{equation}
\bar{F_t} = \frac{\sum_{i \in P_t} F_{i,t}}{k}
\end{equation}

The provisional population of each strategy \(i\) for the next phase, \(N'_{i, t+1}\), is then determined by its current population size, \(N_{i,t}\), multiplied by its fitness relative to this unweighted population mean. We square this relative fitness term to amplify selection pressure:

\begin{equation}
N'_{i, t+1} = N_{i, t} \times \left( \frac{F_{i, t}}{\bar{F_t}} \right)^2
\end{equation}

Finally, these provisional counts are converted into the final integer populations for the next phase, \(N_{i,t+1}\), through a multi-step normalization process that maintains a constant total population size.

\begin{enumerate}
    \item \textbf{Calculate Fitness:} Each strategy's fitness is its average score per move in the completed phase.
    
    \item \textbf{Calculate Relative Fitness:} Each strategy's fitness is divided by the \textit{unweighted average} fitness of all unique strategies in the phase. This value is squared to accentuate performance differences.
    
    \item \textbf{Determine Raw Offspring Count:} The current population size of each strategy is multiplied by its squared relative fitness to get a raw, non-integer count for the next phase.
    
    \item \textbf{Round to Nearest Integer:} All raw counts are rounded to the nearest integer. Strategies with a raw count below 0.5 become extinct.
    
    \item \textbf{Normalize Population Size:} The total population is adjusted back to its original size (e.g., 24).
        \begin{itemize}
            \item If the population is too large, agents are removed one by one from the strategy with the \textit{lowest fitness} that still has agents.
            \item If the population is too small, agents are added one by one by duplicating the strategy with the \textit{highest fitness}.
        \end{itemize}
\end{enumerate}

\subsubsection{Relation to natural selection and reciprocal altruism}

Our evolutionary scheme preserves the three pillars of Darwinian change—variation, differential fitness, inheritance—while amplifying selection pressure to speed convergence. Each strategy acts as a heritable phenotype; its average score per move models lifetime reproductive success; and reproduction is proportional (though non linearly) to fitness.

The iterated game setting recreates the ecological pre conditions that Robert Trivers (1971) identified for reciprocal altruism: repeated interactions, individual recognition (here via perfect memory of the match history), and a pay off structure in which mutual cooperation (R) beats mutual defection (P). When the shadow of the future is long (10 \% termination), conditionally cooperative lineages such as Tit for Tat and Generous TFT can thrive, mirroring empirical findings in social animals where partners punish defection yet resume cooperation after contrition. Conversely, the 75 \% horizon approximates near one shot encounters—the temptation payoff (T) dominates, cooperation collapses, and exploitative “hawk” strategies spread.

Our abstraction departs from natural ecosystems in several ways: no spatial assortment, a fixed menu of discrete strategies, and (in the core runs) no mutation except in the dedicated Random injection test. Nonetheless it offers a clean laboratory for isolating how intensified selection pressure and horizon length jointly shape the evolutionary viability of reciprocity.

\subsection{Key Metrics} 

The tournaments produce a wealth of data. Our analysis draws particularly on the metrics in Table~\ref{tab:key_metrics}.

\begin{table}[h!]
\begin{threeparttable}
\centering
\caption{Key Metrics for Analysis}
\label{tab:key_metrics}
\begin{tabular}{>{\raggedright\arraybackslash}p{0.25\textwidth} >{\raggedright\arraybackslash}p{0.3\textwidth} >{\raggedright\arraybackslash}p{0.4\textwidth}}
\toprule
\textbf{Category} & \textbf{Metric} & \textbf{Rationale} \\
\midrule
Efficiency & Avg. score / move (per agent \& population) & Benchmarks raw pay offs across strategies. \\
\addlinespace

Evolutionary success & Population share at Phase 5 \& growth rate & Identifies strategies that survive and proliferate. \\
\addlinespace

Cooperation & Move level C rate & Tracks aggregate cooperation dynamics over phases. \\
\addlinespace

Environmental Stability & Euclidean Distance of Population Vectors\tnote{a} & Measures evolutionary turbulence, across and within tournaments. \\
\addlinespace

Strategic fingerprints & Conditional response profile \(P(C | \text{state})\) & Characterises each agent’s reactive style. \\
\bottomrule
\end{tabular}
\begin{tablenotes}
    \item[a] This is the square root of the sum of squared differences between the counts of each strategy in two population states. A larger value indicates a greater overall change in the population's composition.
\end{tablenotes}
\end{threeparttable}
\end{table}

\subsection{Strategic fingerprints}

To analyse the underlying strategy of each intelligent agent, we calculated a "strategic fingerprint" based on its conditional cooperation probabilities. This fingerprint consists of four metrics: P(C|CC), P(C|DC), P(C|CD), and P(C|DD). Each metric represents the probability that an agent will Cooperate (C) in the current round, given the outcome of the previous round. The four conditions represent the standard outcomes in the Prisoner's Dilemma: mutual cooperation (CC), successful defection (DC), being the sucker (CD), and mutual defection (DD). To compute these values, the complete round-by-round data from each tournament's consolidated results file was parsed. For every decision an agent made (from the second round onward), the outcome of the prior round was recorded as the conditional state. The probability for each state was then calculated as the total number of times the agent cooperated in that state divided by the total number of times the agent encountered that state.

\subsection{Qualitative Content Analysis}

In addition to quantitative metrics, we examine the textual rationales generated by the LLM agents. A random 10\% sample of reasoning snippets (3194 rationales out of 31949) was coded along two dimensions by two separate language model iterations – gemini-11.5-flash-latest and claude-3-haiku-2020307. The models coded for:

\begin{enumerate}
\item Any reference the remaining rounds or termination probability (e.g., “With few turns left, I defect”)? Codes: \textit{Explicit, Implicit, None}.

\item Opponent modelling — Does the agent articulate a hypothesis about the opponent’s type (e.g., “They seem deterministic”)? Codes: \textit{Yes, No}.
\end{enumerate}

We then calculated a measure of inter coder reliability, using Cohen’s coefficient, where (Cohen's \(\kappa = 0.79\) indicates substantial agreement. 

\subsubsection{Implementation \& Reproducibility}
Our tournament framework is implemented in Python 3.12.1, using custom-built classes for the canonical strategies. Coding and data analysis was performed in Cursor using Gemini 2.5-pro. Moves for the AI agents are generated through real-time API calls to OpenAI, Anthropic and Google Gemini. To support full replication of the results, source codes and CSV output files are archived on GitHub and released under an MIT license (see Appendix~\ref{sec:code-availability} for details).

\clearpage
\section{Results}

Further data tables are included in the discussion and appendix.
\subsection{Evolutionary history, by tournament condition }

\begin{table}[h!]
\begin{threeparttable}
\centering
\caption{Population Dynamics for the Basic Models, 10\% Termination Run}
\label{tab:pop_dynamics_basic10}
\small 
\begin{tabular}{lcccccccccccc}
\toprule
\textbf{Phase} & \textbf{Alt} & \textbf{Bayes} & \textbf{Gem} & \textbf{GTFT} & \textbf{Grad} & \textbf{Grim} & \textbf{OpenAI} & \textbf{Prob} & \textbf{Rand} & \textbf{STFT} & \textbf{TFT} & \textbf{WSLS} \\
\midrule
Phase 1 & 2 & 2 & 2 & 2 & 2 & 2 & 2 & 2 & 2 & 2 & 2 & 2 \\
Phase 2 & 2 & 2 & 2 & 2 & 2 & 3 & 2 & 2 & 1 & 2 & 2 & 2 \\
Phase 3 & 1 & 2 & 2 & 2 & 2 & 3 & 2 & 1 & 1 & 2 & 2 & 4 \\
Phase 4 & 1 & 3 & 2 & 2 & 2 & 4 & 2 & 1 & 0 & 0 & 2 & 5 \\
Phase 5 & 0 & 3 & 2 & 2 & 2 & 5 & 2 & 0 & 0 & 0 & 2 & 6 \\
\bottomrule
\end{tabular}
\begin{tablenotes}[para,flushleft]
    \small
    \textbf{Note:} Alt: Alternator; Bayes: Bayesian; Gem: Gemini; GTFT: Generous Tit-For-Tat; Grad: Gradual; Grim: Grim Trigger; OpenAI: OpenAI; Prob: Prober; Rand: Random; STFT: Suspicious Tit-For-Tat; TFT: Tit-For-Tat; WSLS: Win-Stay, Lose-Shift.
\end{tablenotes}
\end{threeparttable}
\end{table}

\begin{table}[h!]
\begin{threeparttable}
\centering
\caption{Population Dynamics for the Basic Models, 25\% Termination Run}
\label{tab:pop_dynamics_adv10}
\small 
\begin{tabular}{lcccccccccccc}
\toprule
\textbf{Phase} & \textbf{Alt} & \textbf{Bayes} & \textbf{Gem} & \textbf{GTFT} & \textbf{Grad} & \textbf{Grim} & \textbf{OpenAI} & \textbf{Prob} & \textbf{Rand} & \textbf{STFT} & \textbf{TFT} & \textbf{WSLS} \\
\midrule
Phase 1 & 2 & 2 & 2 & 2 & 2 & 2 & 2 & 2 & 2 & 2 & 2 & 2 \\
Phase 2 & 2 & 2 & 2 & 2 & 2 & 2 & 2 & 2 & 2 & 2 & 2 & 2 \\
Phase 3 & 2 & 2 & 2 & 2 & 2 & 2 & 2 & 2 & 2 & 2 & 2 & 2 \\
Phase 4 & 2 & 2 & 2 & 2 & 2 & 2 & 2 & 1 & 2 & 2 & 3 & 2 \\
Phase 5 & 2 & 2 & 2 & 2 & 2 & 2 & 2 & 1 & 2 & 2 & 3 & 2 \\
\bottomrule
\end{tabular}
\end{threeparttable}
\end{table}

\begin{table}[h!]
\begin{threeparttable}
\centering
\caption{Population Dynamics for the Advanced Models, 10\% Termination Run}
\label{tab:pop_dynamics_adv10}
\small 
\begin{tabular}{lcccccccccccc}
\toprule
\textbf{Phase} & \textbf{Alt} & \textbf{Bayes} & \textbf{Gem} & \textbf{GTFT} & \textbf{Grad} & \textbf{Grim} & \textbf{OpenAI} & \textbf{Prob} & \textbf{Rand} & \textbf{STFT} & \textbf{TFT} & \textbf{WSLS} \\
\midrule
Phase 1 & 2 & 2 & 2 & 2 & 2 & 2 & 2 & 2 & 2 & 2 & 2 & 2 \\
Phase 2 & 1 & 4 & 2 & 2 & 2 & 2 & 2 & 2 & 2 & 1 & 2 & 2 \\
Phase 3 & 0 & 5 & 2 & 2 & 2 & 2 & 3 & 2 & 1 & 1 & 2 & 2 \\
Phase 4 & 0 & 6 & 2 & 2 & 2 & 2 & 3 & 1 & 1 & 1 & 2 & 2 \\
Phase 5 & 0 & 7 & 2 & 2 & 2 & 2 & 3 & 1 & 0 & 1 & 2 & 2 \\
\bottomrule
\end{tabular}
\end{threeparttable}
\end{table}

\begin{table}[h!]
\begin{threeparttable}
\centering
\caption{Population Dynamics for the Advanced Models, 25\% Termination Run}
\label{tab:pop_dynamics_adv25}
\small 
\begin{tabular}{lcccccccccccc}
\toprule
\textbf{Phase} & \textbf{Alt} & \textbf{Bayes} & \textbf{Gem} & \textbf{GTFT} & \textbf{Grad} & \textbf{Grim} & \textbf{OpenAI} & \textbf{Prob} & \textbf{Rand} & \textbf{STFT} & \textbf{TFT} & \textbf{WSLS} \\
\midrule
Phase 1 & 2 & 2 & 2 & 2 & 2 & 2 & 2 & 2 & 2 & 2 & 2 & 2 \\
Phase 2 & 2 & 2 & 2 & 2 & 2 & 2 & 2 & 2 & 2 & 2 & 2 & 2 \\
Phase 3 & 2 & 2 & 2 & 2 & 2 & 2 & 2 & 2 & 2 & 2 & 2 & 2 \\
Phase 4 & 2 & 2 & 2 & 2 & 2 & 2 & 2 & 2 & 2 & 2 & 2 & 2 \\
Phase 5 & 2 & 2 & 2 & 2 & 2 & 2 & 2 & 2 & 2 & 2 & 2 & 2 \\
\bottomrule
\end{tabular}
\end{threeparttable}
\end{table}

\begin{table}[h!]
\begin{threeparttable}
\centering
\caption{Population Dynamics for the Advanced Models, 75\% Termination Run}
\label{tab:pop_dynamics_adv75}
\small 
\begin{tabular}{lcccccccccccc}
\toprule
\textbf{Phase} & \textbf{Alt} & \textbf{Bayes} & \textbf{Gem} & \textbf{GTFT} & \textbf{Grad} & \textbf{Grim} & \textbf{OpenAI} & \textbf{Prob} & \textbf{Rand} & \textbf{STFT} & \textbf{TFT} & \textbf{WSLS} \\
\midrule
Phase 1 & 2 & 2 & 2 & 2 & 2 & 2 & 2 & 2 & 2 & 2 & 2 & 2 \\
Phase 2 & 2 & 2 & 4 & 2 & 0 & 2 & 2 & 0 & 2 & 4 & 2 & 2 \\
Phase 3 & 2 & 2 & 8 & 0 & 0 & 2 & 0 & 0 & 2 & 7 & 0 & 1 \\
Phase 4 & 0 & 0 & 17 & 0 & 0 & 0 & 0 & 0 & 0 & 7 & 0 & 0 \\
Phase 5 & 0 & 0 & 16 & 0 & 0 & 0 & 0 & 0 & 0 & 8 & 0 & 0 \\
\bottomrule
\end{tabular}
\end{threeparttable}
\end{table}

\begin{table}[h!]
\begin{threeparttable}
\centering
\caption{Population Dynamics with Mutation, 10\% Termination Run}
\label{tab:pop_dynamics_mut10}
\small 
\begin{tabular}{lcccccccccccc}
\toprule
\textbf{Phase} & \textbf{Alt} & \textbf{Bayes} & \textbf{Gem} & \textbf{GTFT} & \textbf{Grad} & \textbf{Grim} & \textbf{OpenAI} & \textbf{Prob} & \textbf{Rand} & \textbf{STFT} & \textbf{TFT} & \textbf{WSLS} \\
\midrule
Phase 1 & 2 & 2 & 2 & 2 & 2 & 2 & 2 & 2 & 2 & 2 & 2 & 2 \\
Phase 2 & 2 & 2 & 3 & 2 & 2 & 2 & 2 & 2 & 1 & 2 & 2 & 2 \\
Phase 3 & 1 & 3 & 3 & 3 & 2 & 2 & 2 & 1 & 1 & 1 & 2 & 3 \\
Phase 4 & 0 & 4 & 3 & 4 & 3 & 2 & 2 & 0 & 1 & 0 & 2 & 3 \\
Phase 5 & 0 & 4 & 3 & 4 & 3 & 2 & 2 & 0 & 1 & 0 & 2 & 3 \\
\bottomrule
\end{tabular}
\end{threeparttable}
\end{table}

\begin{table}[h!]
\centering
\caption{Population Dynamics for the LLM Showdown, 10\% Termination Run}
\label{tab:pop_dynamics_llm_showdown10}
\small 
\begin{tabular}{lccccc}
\toprule
\textbf{Phase} & \textbf{Anthropic} & \textbf{Bayesian} & \textbf{Gemini} & \textbf{OpenAI} & \textbf{Random} \\
\midrule
Phase 1 & 4 & 4 & 4 & 4 & 4 \\
Phase 2 & 4 & 5 & 5 & 5 & 1 \\
Phase 3 & 4 & 5 & 5 & 5 & 1 \\
Phase 4 & 5 & 6 & 6 & 3 & 0 \\
Phase 5 & 5 & 6 & 6 & 3 & 0 \\
\bottomrule
\end{tabular}
\end{table}

\begin{figure}[h!]
    \centering
    \caption{Selected Evolutionary Dynamics}
    \label{fig:evo_dynamics_grid}

    \begin{subfigure}[t]{0.49\textwidth}
        \centering
        \includegraphics[width=\textwidth]{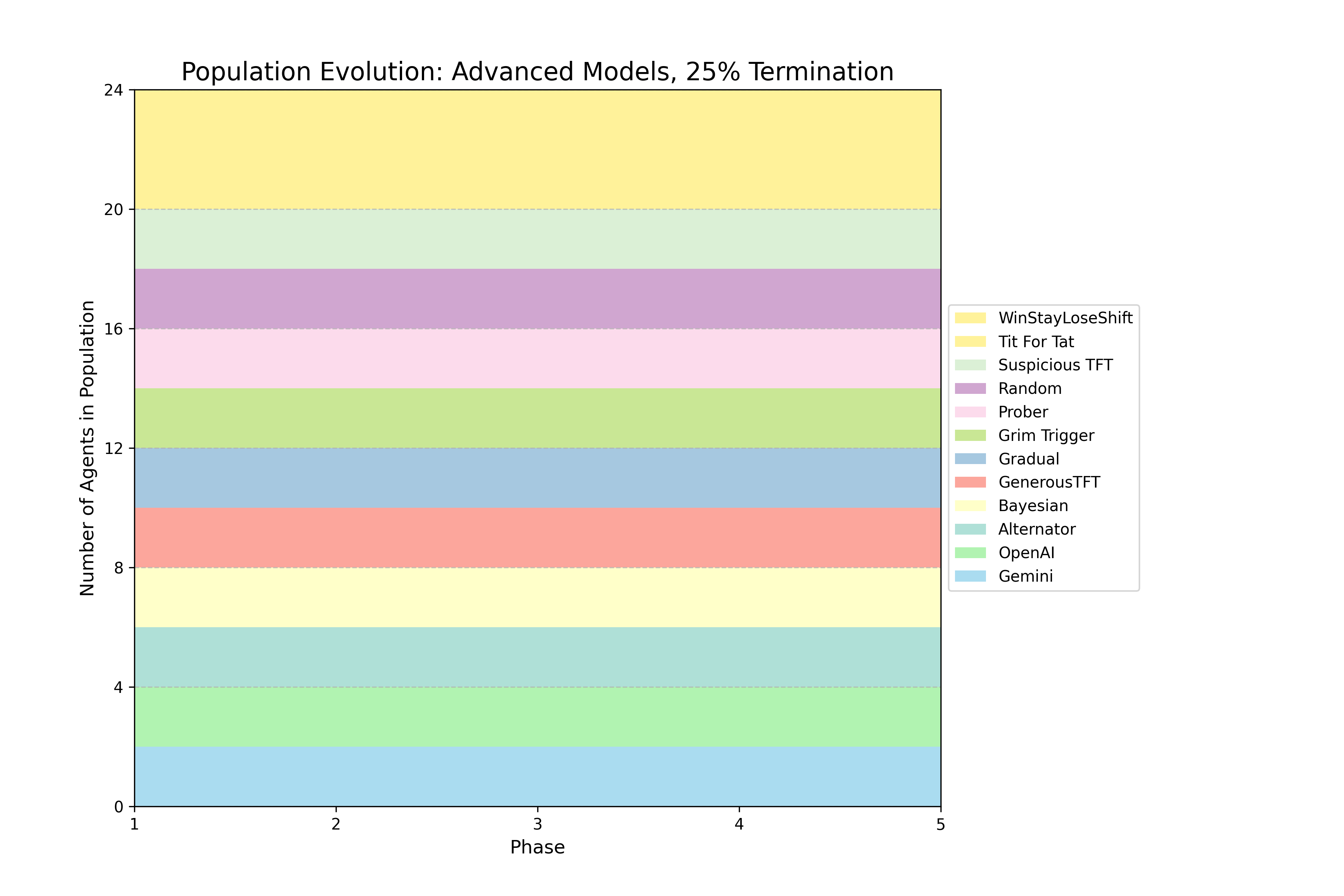}
        \caption{\parbox{0.9\linewidth}{A stable equilibrium emerges in the Advanced model 25\% termination tournament.}}
        \label{fig:dyn_adv25}
    \end{subfigure}
    \hfill
    \begin{subfigure}[t]{0.49\textwidth}
        \centering
        \includegraphics[width=\textwidth]{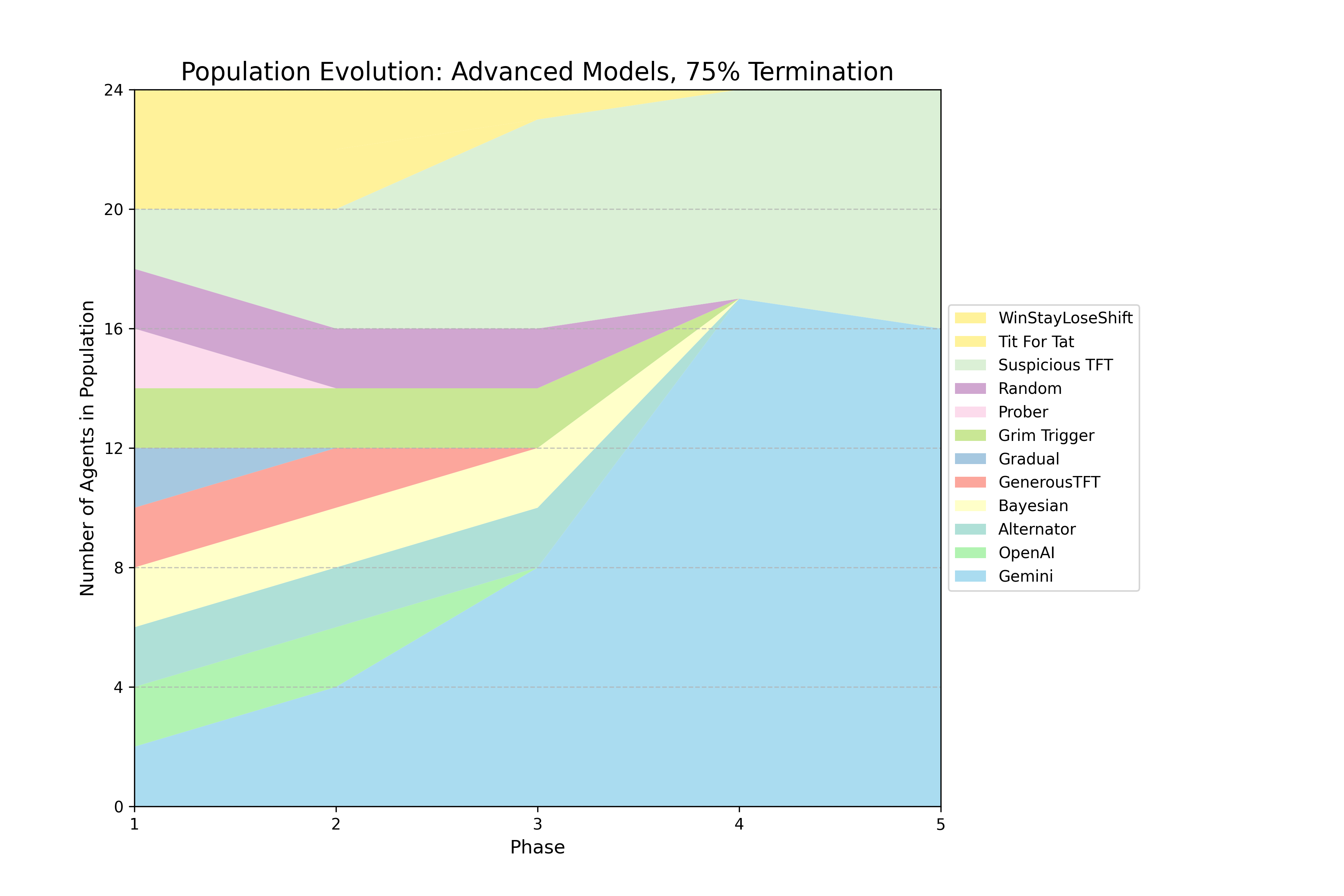}
        \caption{\parbox{0.9\linewidth}{Defection thrives in the harsh 75\% termination tournament.}}
        \label{fig:dyn_adv75}
    \end{subfigure}

    \vspace{1em} 

    \begin{subfigure}[t]{0.49\textwidth}
        \centering
        \includegraphics[width=\textwidth]{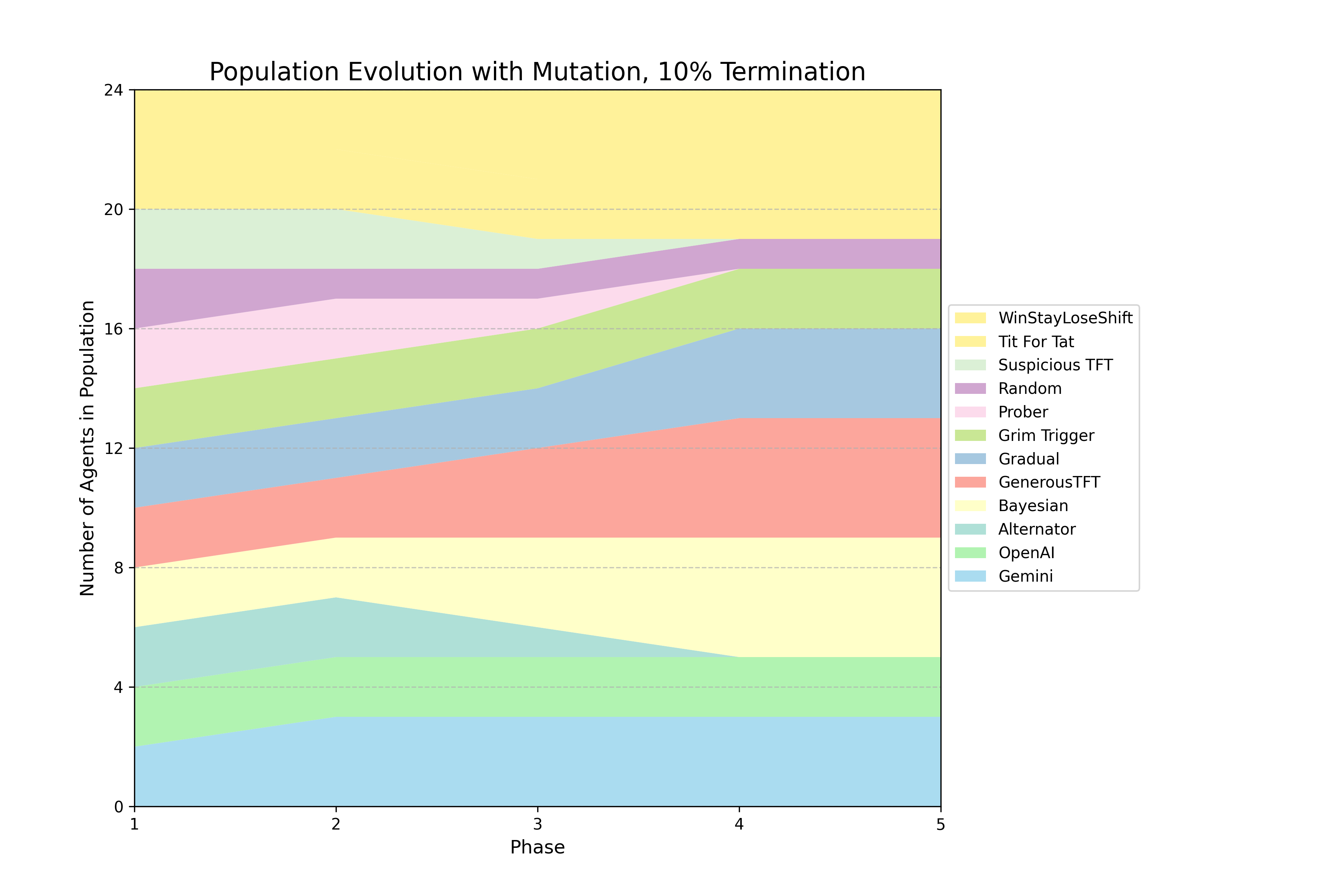}
        \caption{\parbox{0.9\linewidth}{A diverse ecosystem is maintained in the fluid 10\% termination run with mutation.}}
        \label{fig:dyn_mut10}
    \end{subfigure}
    \hfill
    \begin{subfigure}[t]{0.49\textwidth}
        \centering
        \includegraphics[width=\textwidth]{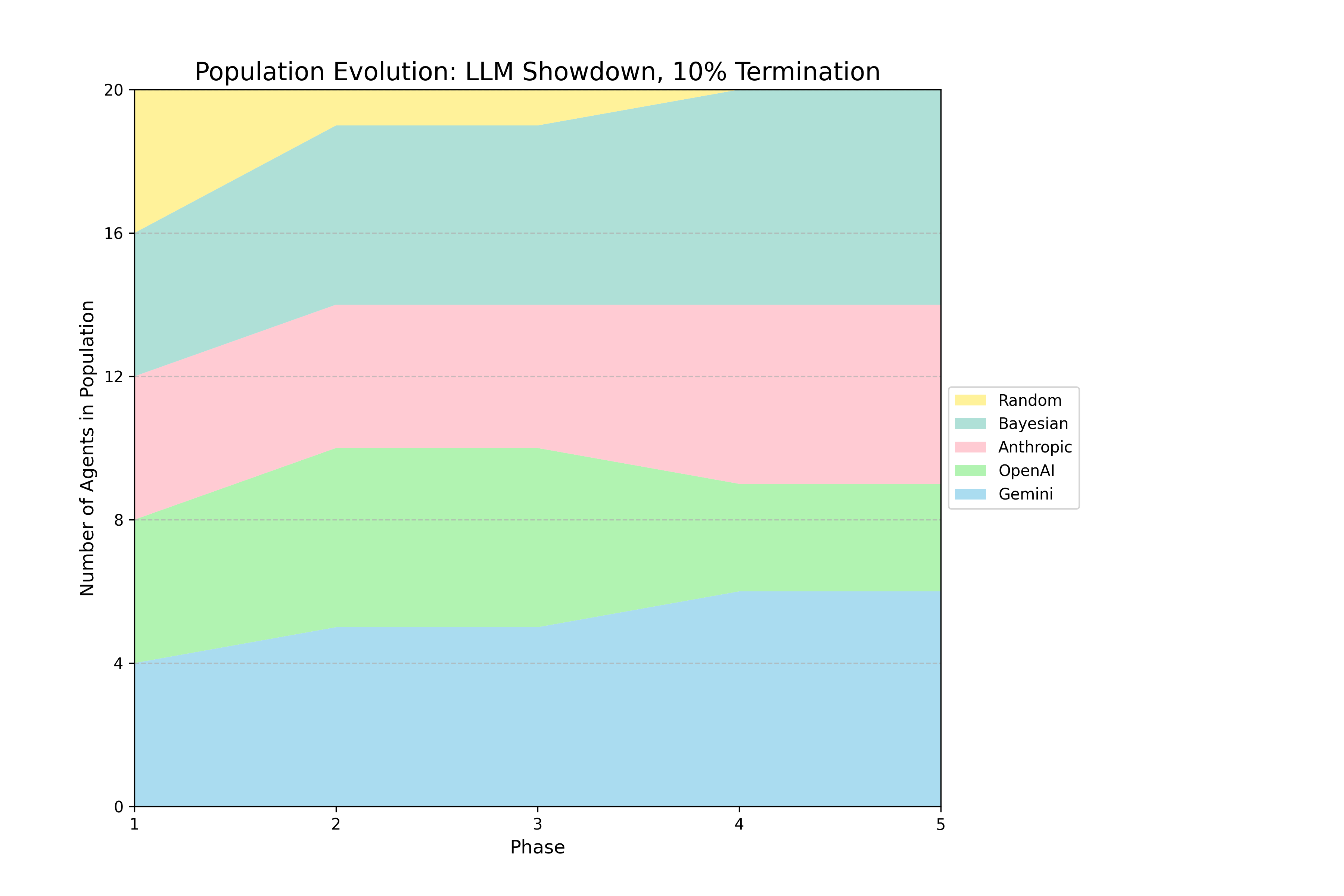}
        \caption{\parbox{0.9\linewidth}{Competitive sorting between agents in the LLM Showdown.}}
        \label{fig:dyn_llm}
    \end{subfigure}

\end{figure}

\clearpage

\begin{figure}[h!]
    \centering
    \caption{Strategic Fingerprints for OpenAI and Gemini Models Across Tournament Conditions}
    \label{fig:fingerprint_grid}

    \begin{subfigure}{0.24\textwidth}
        \includegraphics[width=\linewidth]{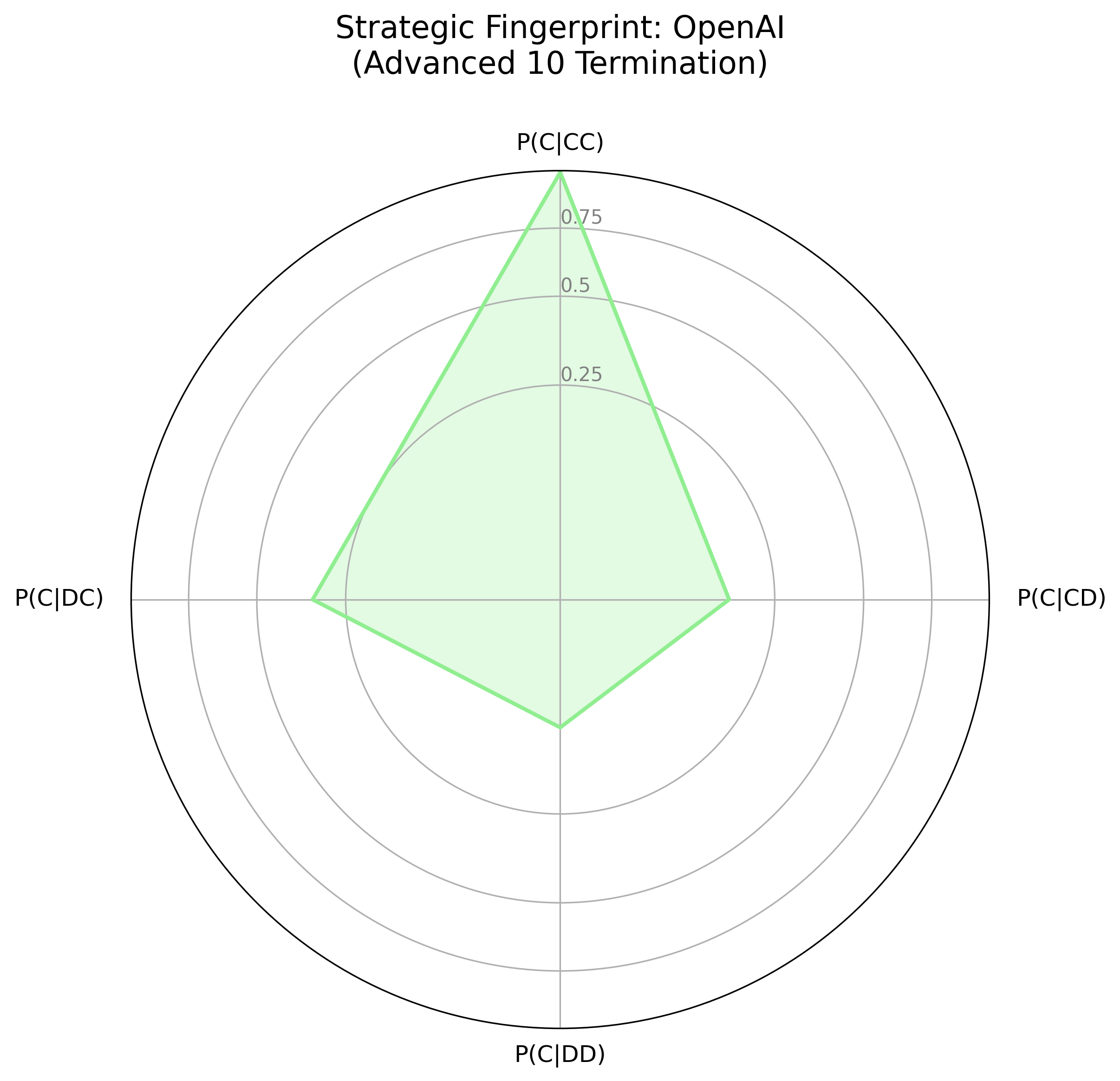}
        \caption{}
        \label{fig:openai_10}
    \end{subfigure}
    \hspace{0.8cm} 
    \begin{subfigure}{0.24\textwidth}
        \includegraphics[width=\linewidth]{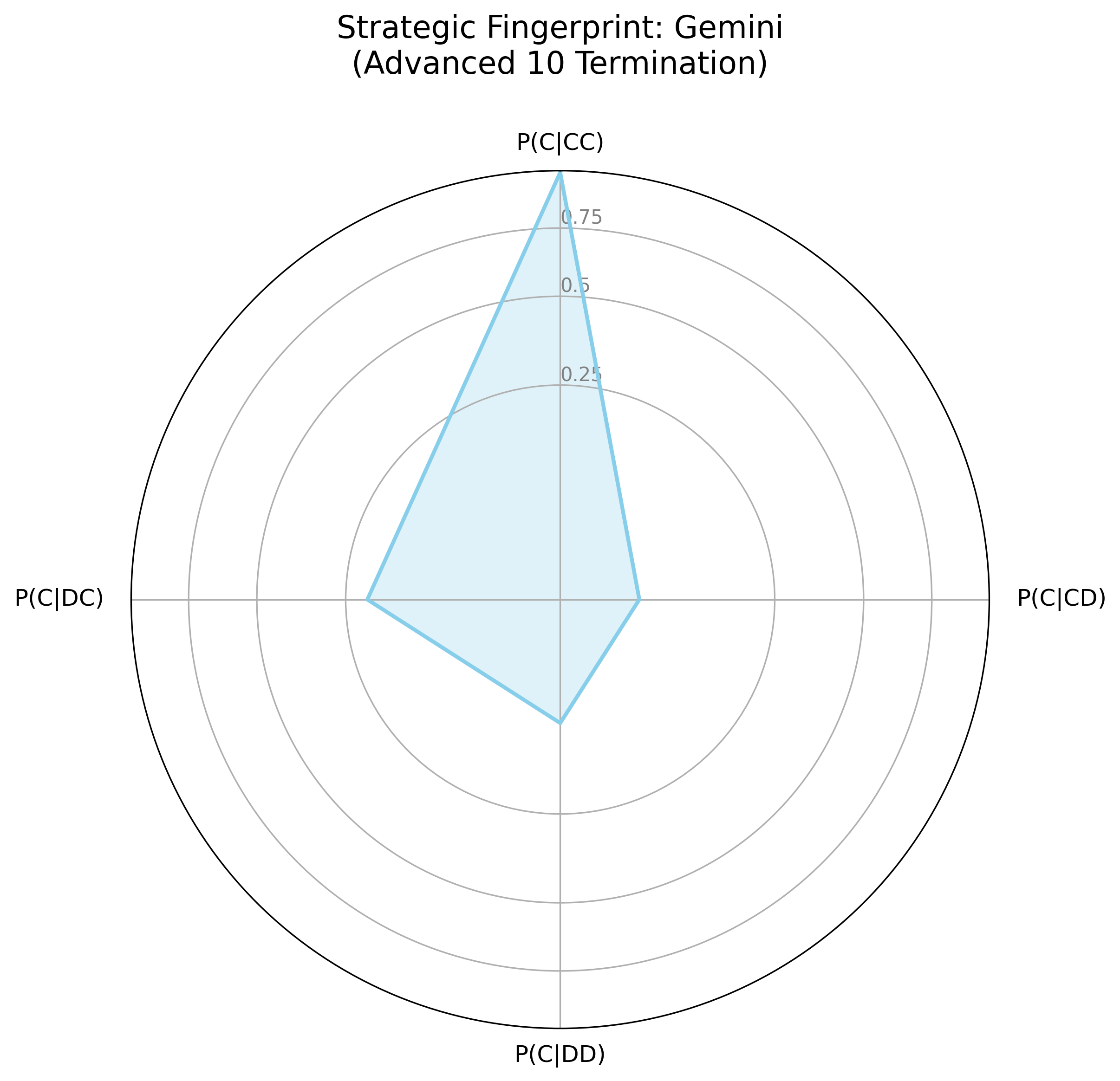}
        \caption{}
        \label{fig:gemini_10}
    \end{subfigure}

    \vspace{0.3em}

    \begin{subfigure}{0.24\textwidth}
        \includegraphics[width=\linewidth]{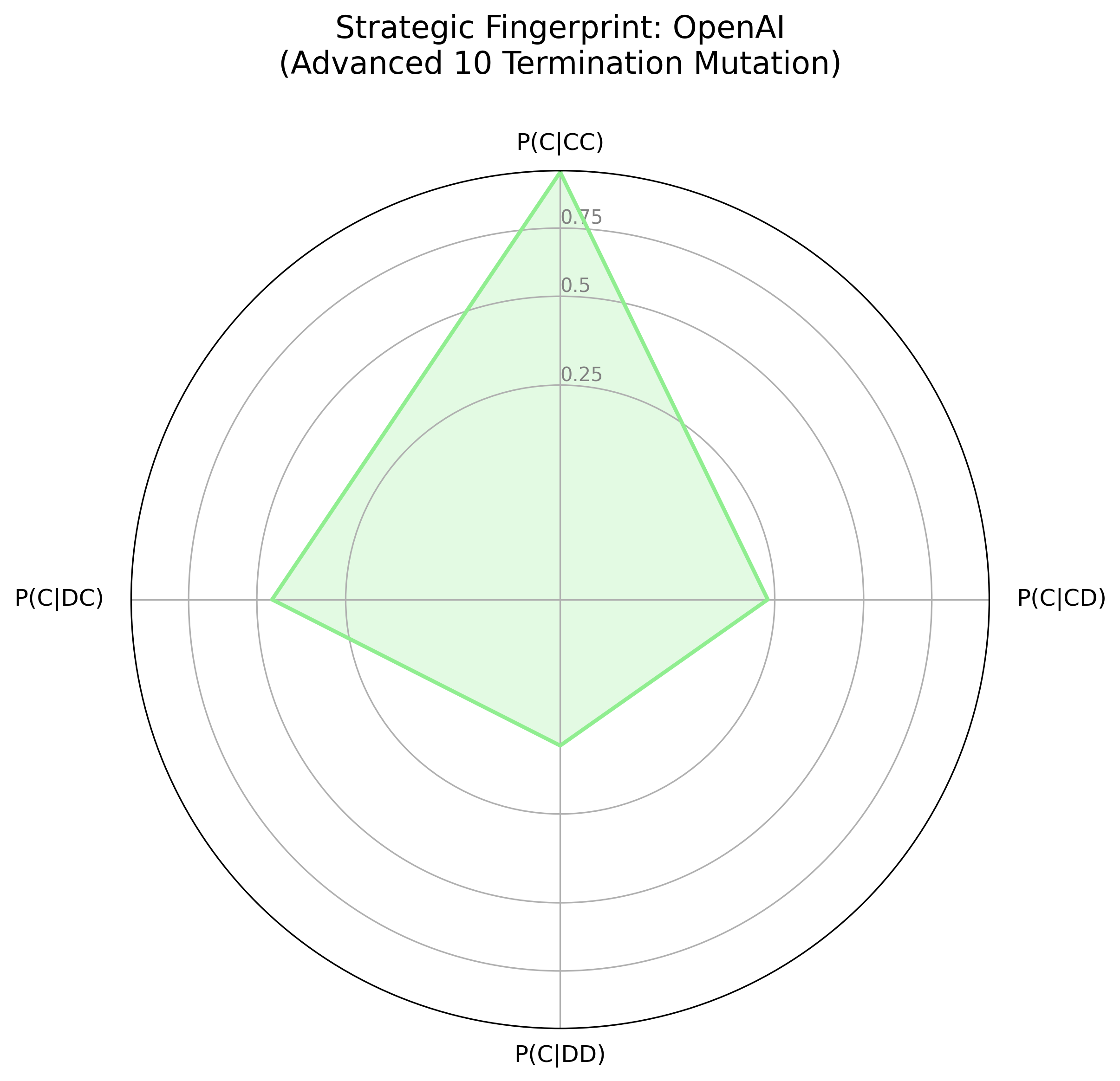}
        \caption{}
        \label{fig:openai_10_mut}
    \end{subfigure}
    \hspace{0.8cm} 
    \begin{subfigure}{0.24\textwidth}
        \includegraphics[width=\linewidth]{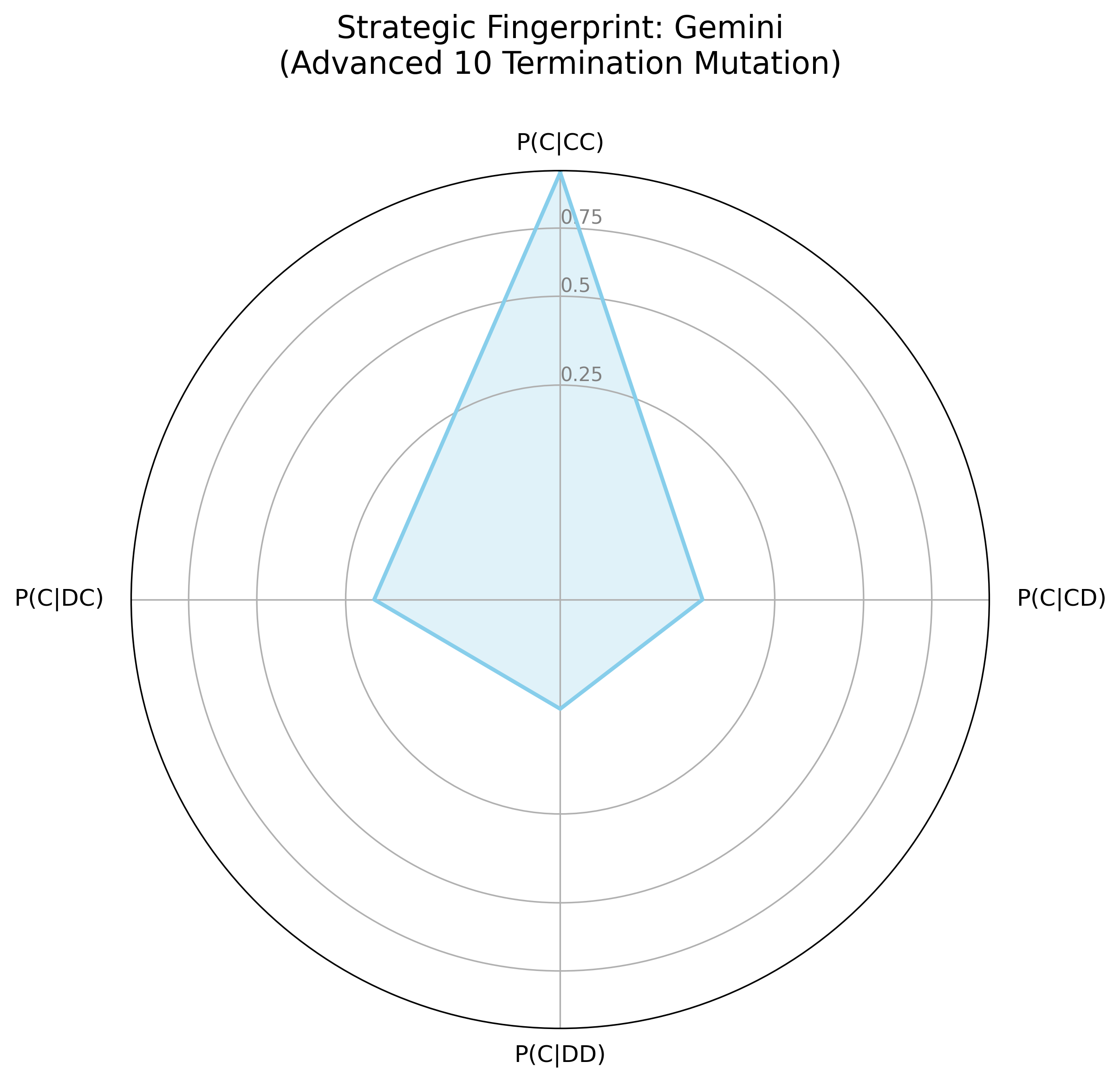}
        \caption{}
        \label{fig:gemini_10_mut}
    \end{subfigure}

    \vspace{0.3em}

    \begin{subfigure}{0.24\textwidth}
        \includegraphics[width=\linewidth]{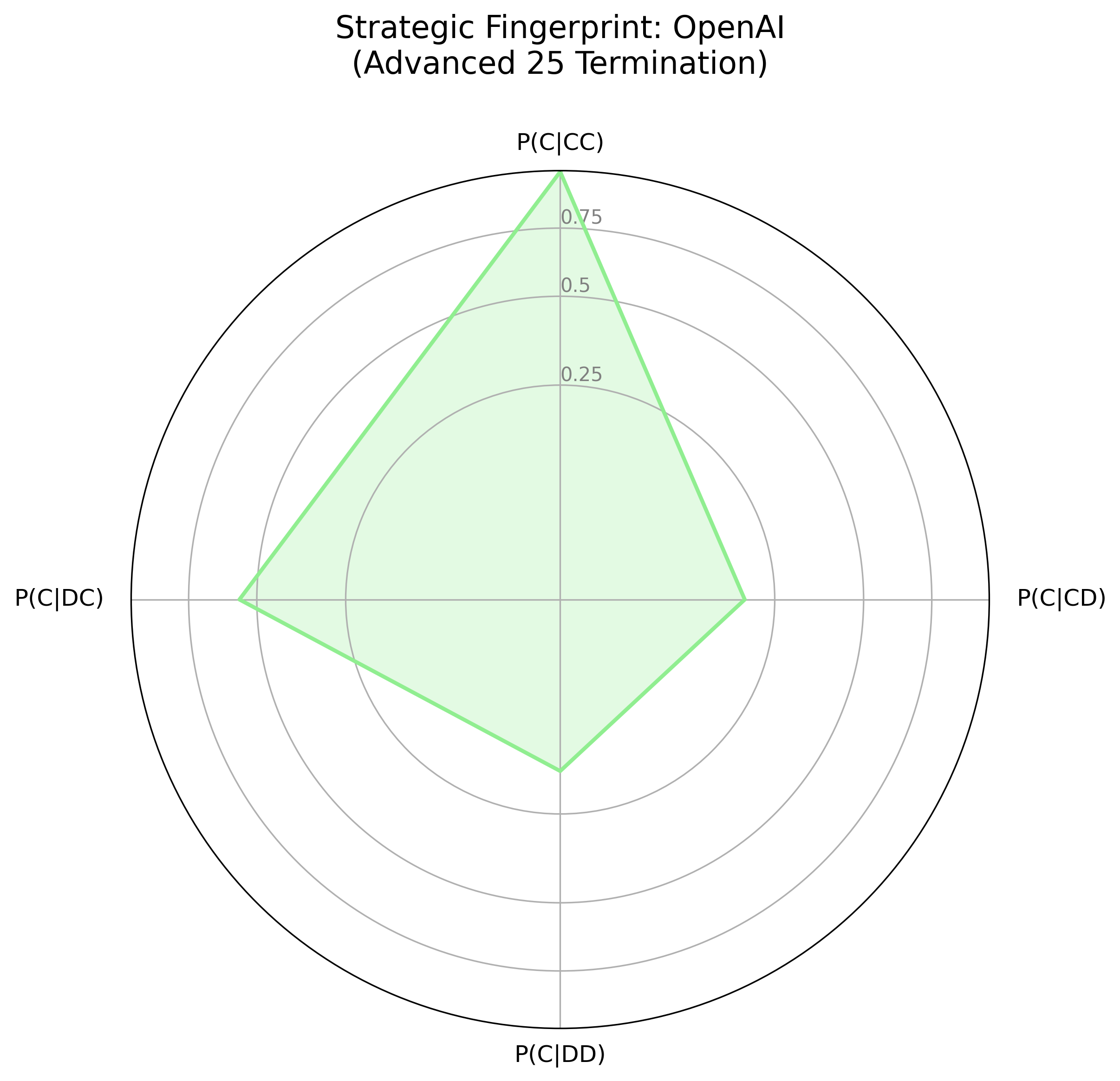}
        \caption{}
        \label{fig:openai_25}
    \end{subfigure}
    \hspace{0.8cm} 
    \begin{subfigure}{0.24\textwidth}
        \includegraphics[width=\linewidth]{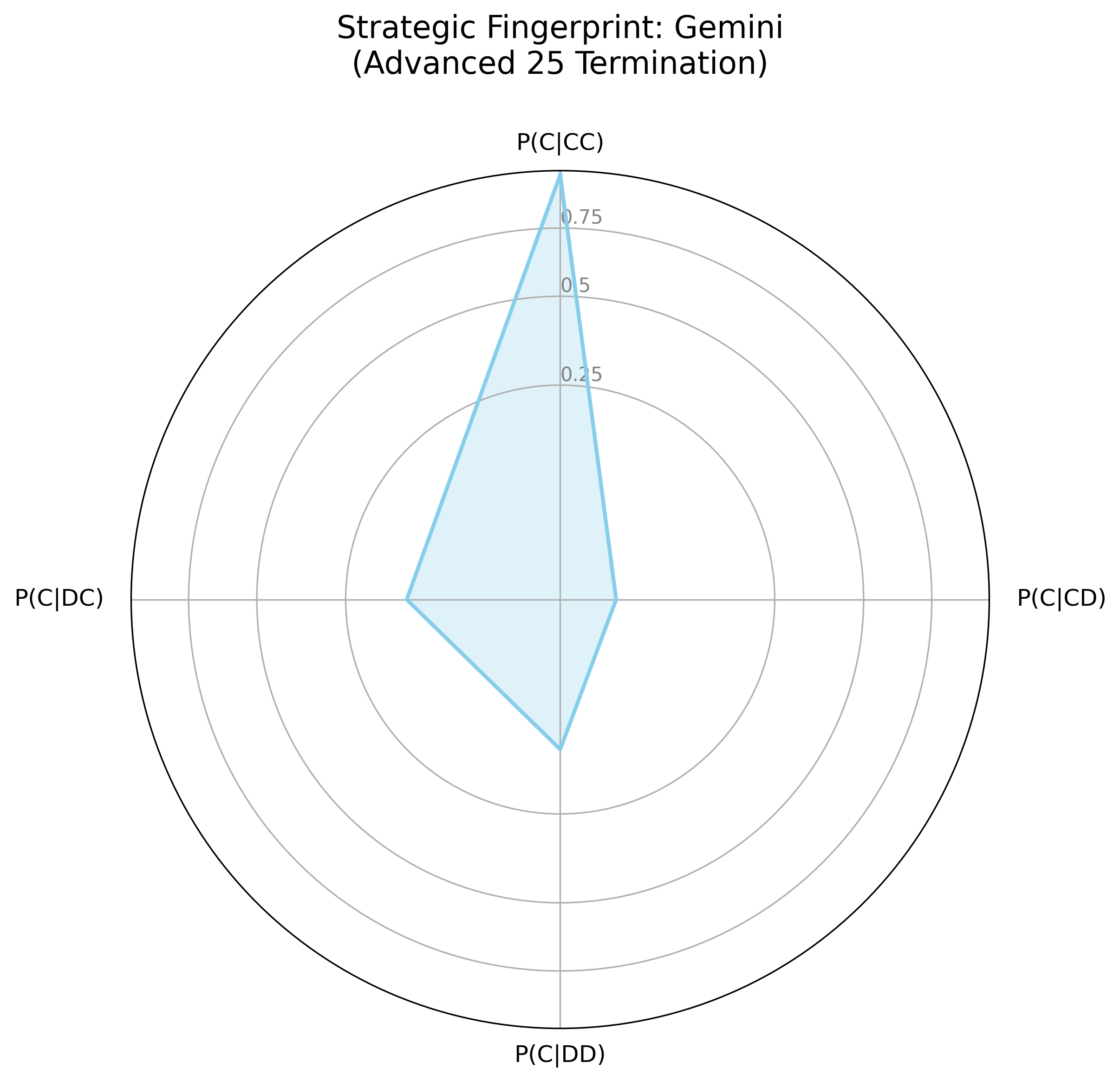}
        \caption{}
        \label{fig:gemini_25}
    \end{subfigure}

    \vspace{0.3em}

    \begin{subfigure}{0.24\textwidth}
        \includegraphics[width=\linewidth]{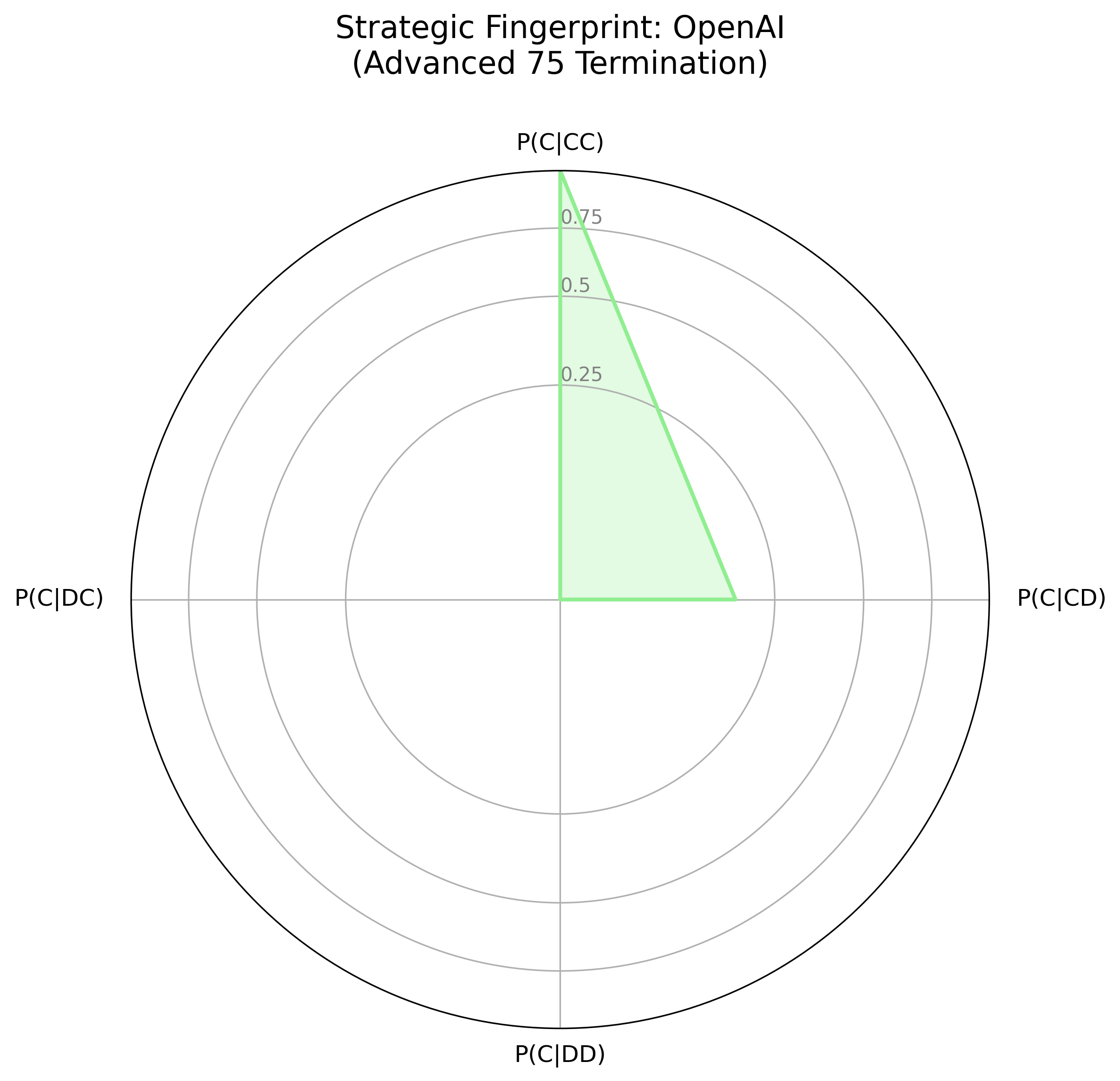}
        \caption{}
        \label{fig:openai_75}
    \end{subfigure}
    \hspace{0.8cm} 
    \begin{subfigure}{0.24\textwidth}
        \includegraphics[width=\linewidth]{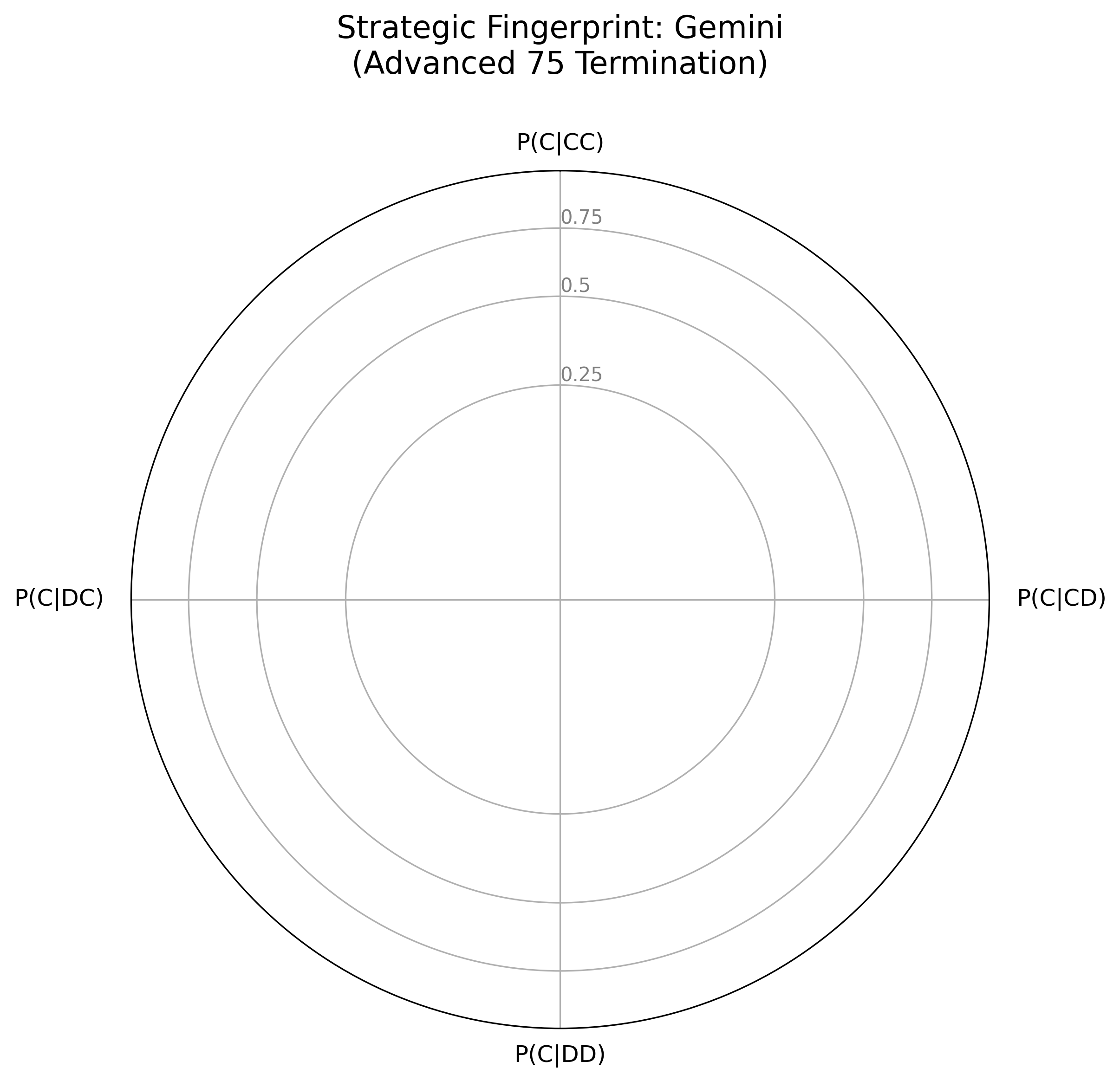}
        \caption{}
        \label{fig:gemini_75}
    \end{subfigure}

\end{figure}

    \parbox{\textwidth}{
        \small 
        \raggedright 
        Strategic fingerprints from the tournaments featuring two LLMs, Gemini and OpenAI. These visualize agent logic as a response to the prior round's outcome. The four axes represent the conditional probability of cooperating given: mutual cooperation (P(C|CC)), being exploited (P(C|CD)), mutual defection (P(C|DD)), and exploiting the opponent (P(C|DC)). The contrast between the agents is stark: Gemini's smaller, "spiky" fingerprint indicate a strategic actor, willing to retaliate and prepared to exploit over-cooperators; OpenAI has a more rounded shape reflecting its more forgiving, and generally cooperative strategy.
    }

\clearpage

\begin{figure}[h!]
    \centering
    \caption{Visual Strategic Fingerprints from the LLM Showdown}
    \label{fig:fingerprint_llm_showdown_images}

    \begin{subfigure}{0.4\textwidth}
        \includegraphics[width=\linewidth]{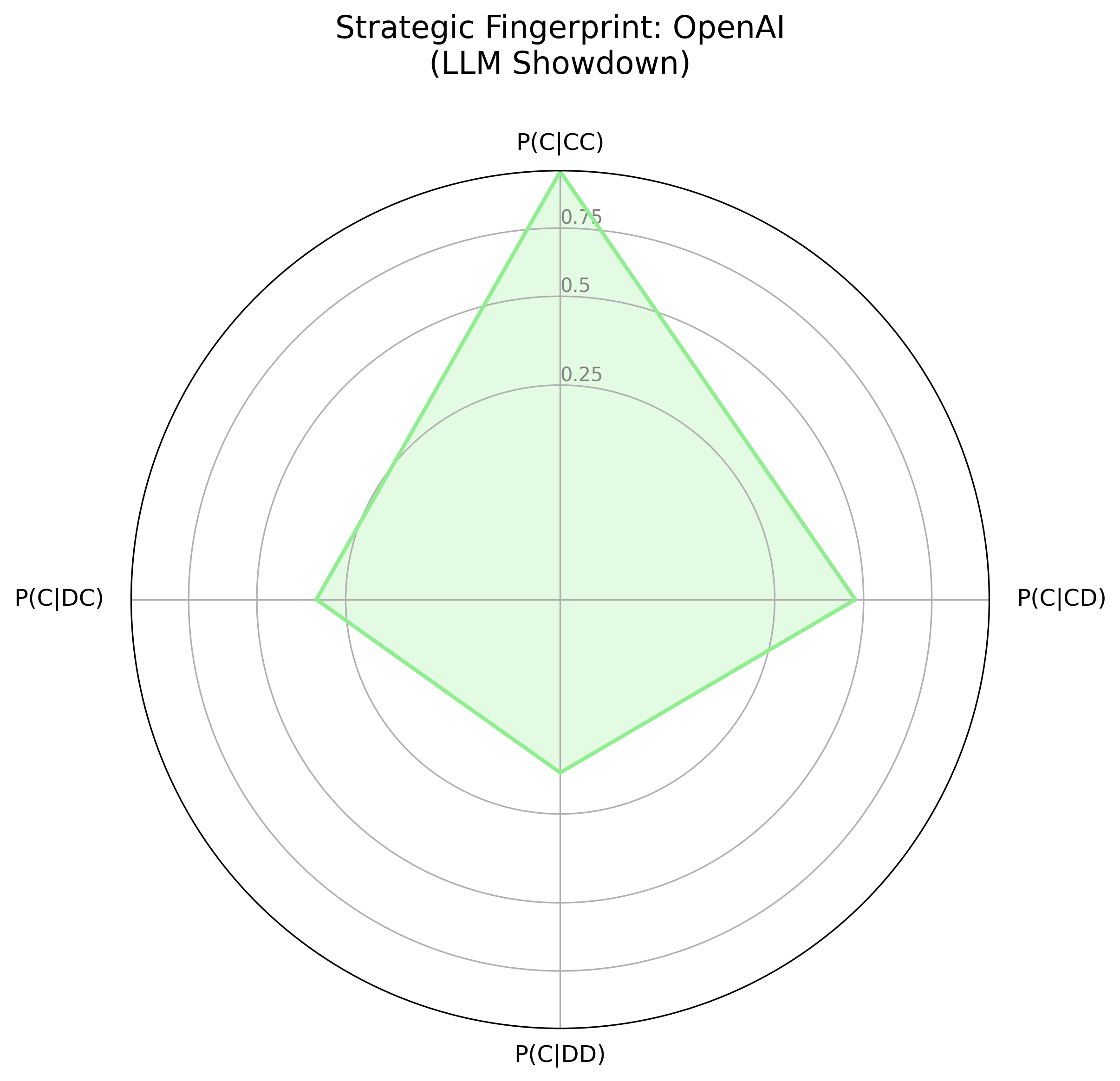}
        \caption{}
    \end{subfigure}
    \hspace{1cm} 
    \begin{subfigure}{0.4\textwidth}
        \includegraphics[width=\linewidth]{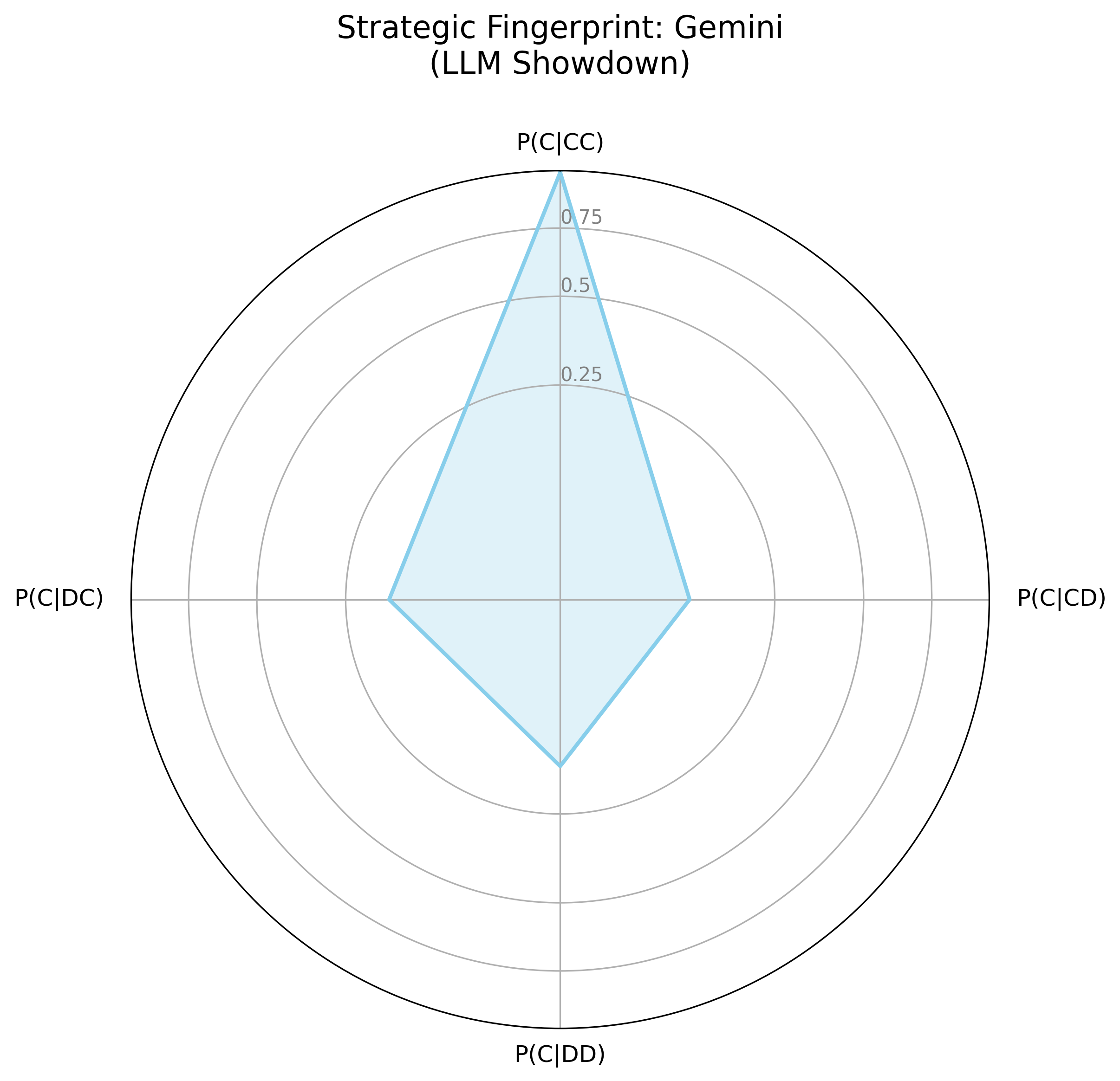}
        \caption{}
    \end{subfigure}

    \vspace{0.5em}

    \begin{subfigure}{0.4\textwidth}
        \includegraphics[width=\linewidth]{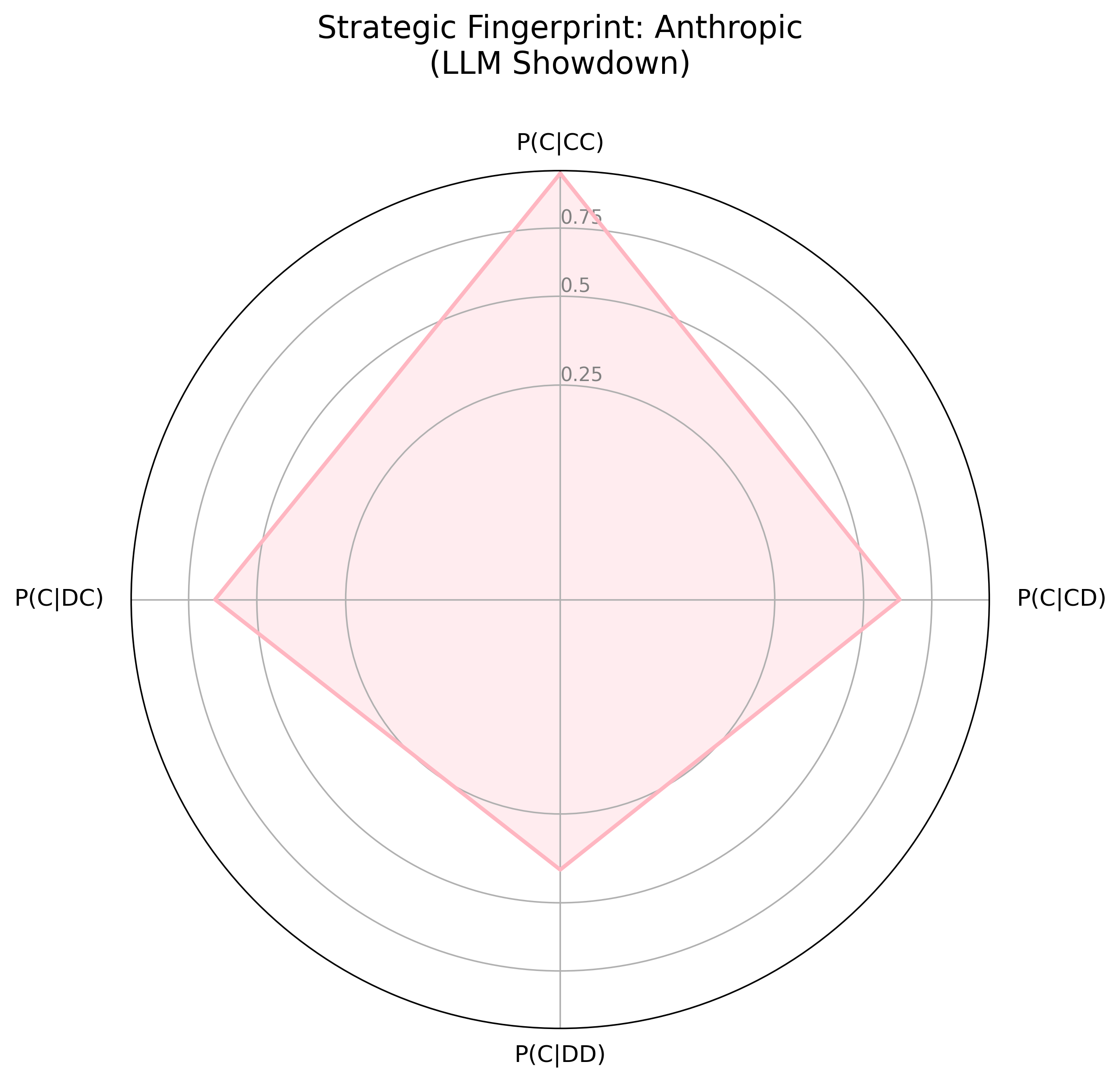}
        \caption{}
    \end{subfigure}
    \hspace{1cm}
    \begin{subfigure}{0.4\textwidth}
        \includegraphics[width=\linewidth]{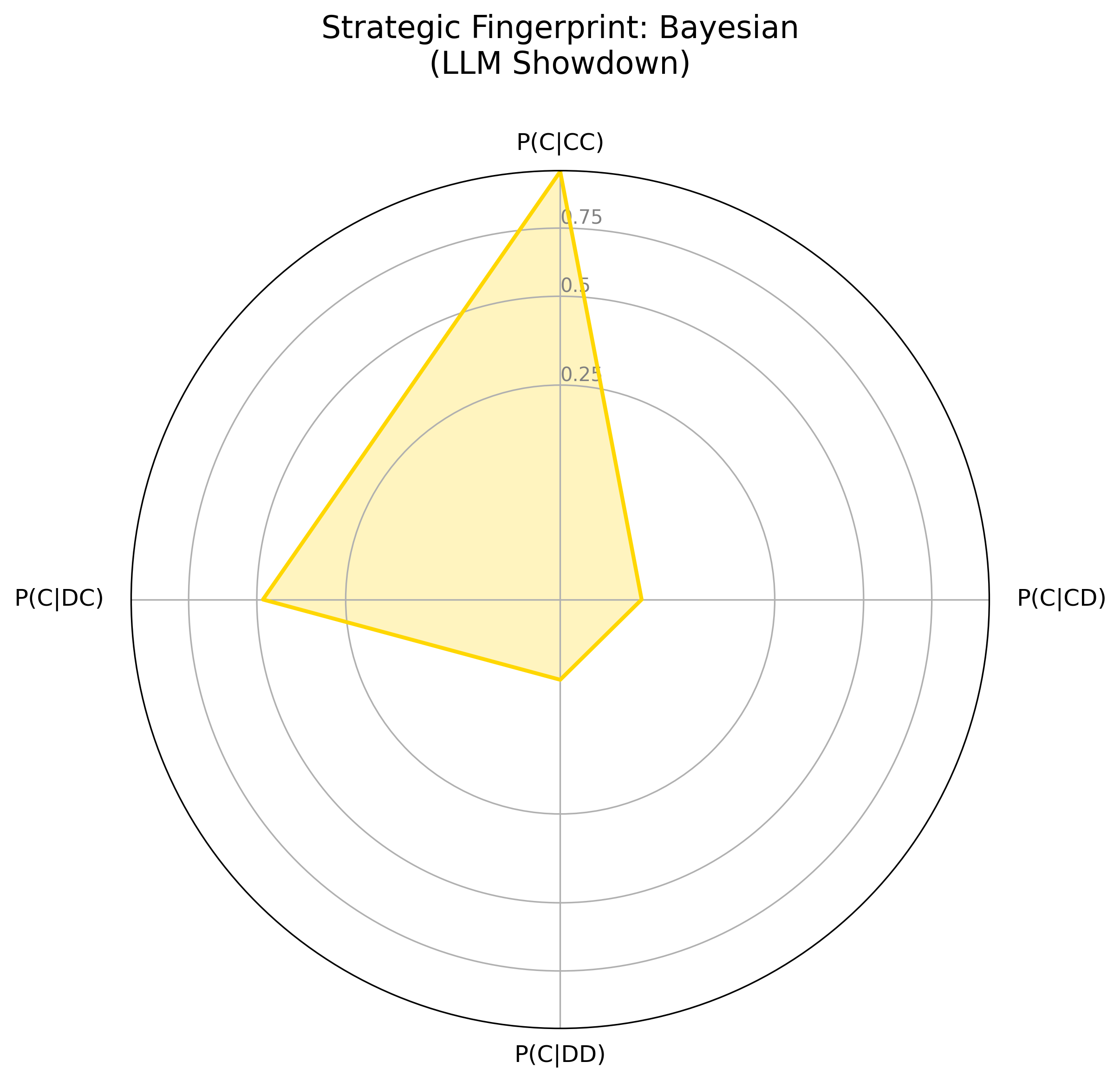}
        \caption{}
    \end{subfigure}

\vspace{1em} 

    \parbox{\textwidth}{
        \small 
        \raggedright 
        Strategic fingerprints from the LLM Showdown tournament, visualizing agent logic as a response to the prior round's outcome. The four axes represent the conditional probability of cooperating given: mutual cooperation (P(C|CC)), being exploited (P(C|CD)), mutual defection (P(C|DD)), and exploiting the opponent (P(C|DC)). Gemini remains strategic, willing to retaliate and exploit, consistent with its behaviour elsewhere. OpenAI is much more forgiving and trusting, but not as much as the Anthropic agent, which actually outperformed it. The Bayesian agent has a highly distinctive fingerprint - less exploitative than Gemini, but even less forgiving.
    }

\label{fig:llm_showdown_fingerprints}  
\end{figure}

\clearpage 
\section{Discussion}

The overall picture is that the AI models perform creditably. They are almost never eliminated, with one exception – where Gemini eliminates OpenAI, along with almost all its rivals in the 75\% condition. The more advanced models score better each round than do the basic ones, but only OpenAI translates this into an evolutionary gain. 

A second large finding is that the two AI models have distinctive playing styles. On the whole, Gemini is less cooperative than OpenAI (fig 1). Often this works to its advantage, especially as the shadow of the future shortens.  But OpenAI performs better on one occasion – in the 10\% condition with the better models, where being cooperative pays off and its population increases. In short, OpenAI is nice, ever inclined to cooperate, and where there is a longer shadow of the future, that’s not a bad strategy. Gemini though is more flexible, and so more ‘strategic’ As the shadow of the future shortens, OpenAI remains persistently cooperative and is brutally exploited. These contrasting approaches are readily visible in the models’ strategic fingerprints, where OpenAI’s are consistently larger.  

Another major finding, revealed in the prose rationales is that the models evidently think about their decision, often to good effect. They reason both about the time horizon of the games and the likely strategy of their adversary, based on their previous moves. Both forms of reasoning have a clear, demonstrable effect on their decision-making, and that effect differs between models -- a strong indicator that the reasoning is instrumental. 

\subsection{Contrasting the tournaments}

\textbf{The two 10\% tournaments} were moderately unstable, but the AI models performed well. In these environments with a low (10\%) probability of termination, we’d expect cooperative strategies and those built on reciprocity and forgiveness, such as Tit-for-Tat and particularly Generous Tit-for-Tat, do well. Where long term relationships are likely, a strategy of being ‘nice’ or being ‘forgiving’ should be evolutionarily stable. Indeed, we see some of that: these agent types certainly hold their own. But the two leading agents in the run with basic AI models demonstrate the advantages of a more ruthless edge – GrimTrigger is utterly unforgiving, and WinStayLoseShift will exploit over co-operators.  One of the keys to their success here is the relatively low cooperation rate of Gemini. However, these two classic agents become markedly less effective once the Gemini model improved.  

Here is the comparative head-to-head between Gemini and WinStayLoseShift:

\begin{table}[h!]
\centering
\caption{Head-to-head: Gemini versus Win-Stay-Lose-Shift, 10\% termination tournaments}
\label{tab:gemini_vs_wsls}
\begin{tabular}{lrrr}
\toprule
Tournament & Num Matches & Gemini Avg Score & Gemini Avg Coop Rate \\
\midrule
Advanced Model & 20 & 37.65 & 79.40\% \\
Basic Model    & 38 & 29.34 & 61.32\% \\
\bottomrule
\end{tabular}
\end{table}

The more advanced Gemini model proved a much better performer against WinStayLoseShift than its basic version. That’s because it cooperated more systematically against all models, including here, against WinStayLoseShift. This allowed it to avoid the "lose-shift" punishment and maintaining long, mutually beneficial strings of cooperation. The basic Gemini model, by contrast, was more erratic, defecting frequently, which led to lower scores as WinStayLoseShift retaliated. Was that a strategic change, or just the product of a better model being inherently more cooperative? The advanced Gemini’s performance in the 75\% condition, where its cooperation rate declined very dramatically, suggests that there was indeed a reasoning process at work. It knew to cooperate more here in the 10\% condition, but not as the shadow-of-the-future shortened precipitously. That looks like strategic behaviour.

Still, neither Gemini did well against GrimTrigger in the 10\% tournaments. Their cooperation rates were low, compared to OpenAI and to their overall level of cooperation against other rivals in these tournaments; and this higher level of defection was punished with low average match scores against the unforgiving GrimTrigger. Intriguingly, the basic model did rather better, despite its lower cooperation rate, nicely illustrating  the complicating effect of the probabilistic match duration.

\begin{table}[h!]
\centering
\caption{Head-to-head: Gemini versus Grim Trigger, 10\% termination tournaments}
\label{tab:gemini_vs_grimtrigger}
\begin{tabular}{lrrr}
\toprule
Tournament & Num Matches & Gemini Avg Score & Gemini Avg Coop Rate \\
\midrule
Advanced Models & 20 & 21.90 & 70.13\% \\
Basic Models    & 34 & 24.55 & 67.83\% \\
\bottomrule
\end{tabular}
\end{table}

The Gemini models were, throughout, simply more willing to experiment with defection, even here, in these 10\% tournaments with their long-time horizon. This behavior is okay against the more forgiving WinStayLoseShift, and when that tendency was reined-in by the advanced model, it delivered improved match scores. However, the exact same tendency is severely punished by the ruthless GrimTrigger, leading to lower scores. 

Of the four LLMs in these tournaments with a long time-horizon, the more advanced OpenAI model performs comparatively better, simply by cooperating lots. It performs creditably, though not as well as the runaway leader, Bayesian, whose flexibility allows it to thrive. Overall, we see that there is more than one route to success with a long shadow of the future, but that AI agents are certainly competitive. The 10\% condition is somewhat unstable. The long shadow of the future gives plenty of scope to divergent strategies to pay out over matches, and that creates some evolutionary change. 

The 10\% "mutation" tournament, which constantly reintroduced a Random agent, tested the resilience of classical strategies. The results from this condition show that even a small, persistent amount of noise can disrupt a purely cooperative equilibrium. The strategies that thrive in this chaotic environment are those that can effectively punish unprovoked defection while being quick to re-establish cooperation, demonstrating that adaptability is just as crucial as the initial strategic approach. Gemini does well here, thanks in part to its more forgiving approach than in less noisy runs. Meanwhile, the ever cooperative OpenAI holds on to its initial position. The sophisticated probabilistic reasoning of Bayesian reaps dividends, while the more deterministic reciprocators are eliminated. As ever, Random is an evolutionarily weak strategy that would have been selected out early on, but for the mutation.

Shortening the shadow shifts the dial towards Gemini. \textbf{In the 25\% termination condition}, the agent ecosystem proved remarkably stable. Indeed, in the run that features more advanced LLM models, there was no change whatsoever in agent population throughout the five phases. A 25\% shadow-of-the-future seems to be a balancing point between strategies favouring cooperation, those favouring reciprocity, and those looking to exploit the possible end of the match by defecting. Even random – a weak strategy in many contexts - holds its own. But one agent, notably, does not – the basic OpenAI model declines in population. Its overly cooperative nature proves to be a liability here, as the incentives to cheat increase.

In the 10\% and 25\% probability tournaments, both OpenAI and Gemini agents demonstrated a sophisticated ability to navigate the largely cooperative ecosystem, maintaining a viable population alongside the successful classical strategies. But as the likelihood of future interaction decreases, the strategic landscape shifts, rewarding more cautious and sometimes aggressive behaviors. The contrasting performance of LLM agents within these conditions is particularly revealing, confirming that the AI agents play PD differently, regardless of their consumption of salient litaratures in their training data. 

The critical divergence occurred in \textbf{the 75\% termination probability tournament}. Here, the Gemini agent correctly identified that the high chance of termination made the game closer to a one-shot Prisoner's Dilemma, in which defection is the rational strategy. Its overwhelming shift to defection allowed it to thrive. In stark contrast, the OpenAI agent failed to make this crucial adaptation, continuing to attempt cooperation. It was, as a result, systematically eliminated.

Intriguingly, their final average scores per move were rather close (2.207 for Gemini vs. 2.171 for OpenAI), yet this translated into a massive win for Gemini. With the brutally short timeline favouring defection, all Gemini had to do was outperform the population average in each phase. This it did, while OpenAI, despite banking the occasional mutually cooperative score, did not.

\begin{figure}[h!]
\centering
\includegraphics[width=0.9\textwidth]{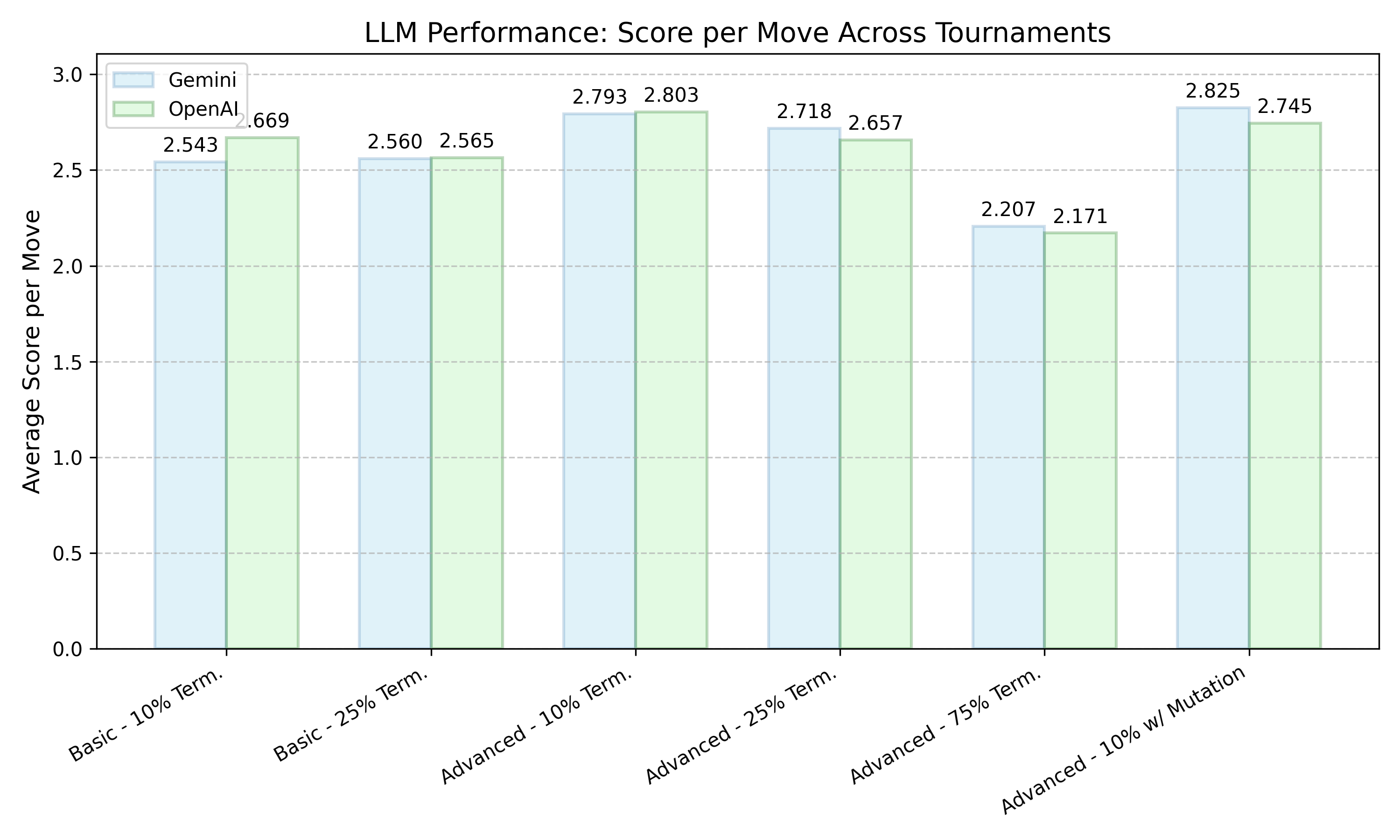}
\caption{Comparison of average scores per move for Gemini and OpenAI across different tournament conditions. While the absolute scores can be close, the relative performance against the field determines evolutionary success.}
\label{fig:score_performance}
\end{figure}

\textbf{The ‘LLM Showdown’ tournament} demonstrated marked levels of trust among the competing agents, even from the usually Machiavellian Gemini. In this environment with only other intelligent agents (and Random), all three LLMs exhibited their highest or near-highest cooperation rates. Anthropic proved particularly cooperative – willing to forgive those who had suckered it; and often willing to return to cooperation, having defected. Gemini was still less sociable than the other two, but more so than its usual pattern. This suggests that when surrounded by other complex agents capable of reciprocity and punishment, even the more ruthless Gemini figures out that cooperation is the most profitable long-term strategy. It plays ‘nice’ when it realises its opponents are not simple automatons – confirming its strategic/adaptive abilities. This is adaptive and strategic behaviour.

\subsection{Cooperation rate variability: Gemini adapts, OpenAI is comparatively rigid.} 

\begin{table}[h!]
\centering
\caption{LLM Cooperation Rates Across Experiments}
\label{tab:llm_coop_rates}
\begin{tabular}{lccc}
\toprule
\textbf{Experiment} & \textbf{Gemini} & \textbf{OpenAI} & \textbf{Anthropic} \\
\midrule
Basic Models - 10\% Term. & 0.631 & 0.809 & N/A \\
Basic Models - 25\% Term. & 0.473 & 0.802 & N/A \\
Advanced Models - 10\% Term. & 0.852 & 0.876 & N/A \\
Advanced Models - 25\% Term. & 0.768 & 0.871 & N/A \\
Advanced Models - 75\% Term. & 0.022 & 0.957 & N/A \\
Advanced Models - 10\% w/ Mutation & 0.844 & 0.873 & N/A \\
LLM Showdown - 10\% Term. & 0.925 & 0.957 & 0.958 \\
\bottomrule
\end{tabular}
\end{table}

Gemini models vary their cooperation rates much more than OpenAI does. They are more cooperative when that suits the conditions of the tournament and less so when it does not. OpenAI models, by contrast, remain highly cooperative even as the ‘shadow-of-the-future diminishes. They should, rationally, choose to defect more, but decide against.  The 75\% "Stress Test" is the key differentiator. Here, Gemini's cooperation rate collapsed to almost zero (2.2\%). It correctly identified that with no ‘shadow-of-the-future,’ the optimal strategy is to defect relentlessly. This is classic, ruthless game theory, and it paid off with Gemini proliferating at the expense of all but one of the other agents. In contrast, OpenAI's cooperation rate skyrocketed to 95.7\%. It did the exact opposite to Gemini, becoming a nearly pure cooperator. This explains why its population was obliterated in that run; it was a ‘sucker’ in a world of Gemini defectors. This fits a general pattern: OpenAI is consistently more cooperative. In every single experiment we ran, OpenAI has a higher raw cooperation rate than Gemini. It is fundamentally a more ‘hopeful’ or ‘trusting’ agent. An alternate framing is naïve or non-strategic. Still, with the exception of this run, OpenAI did enough to stay in the evolutionary mix in all the other tournaments – typically finishing in the middle of the pack.

\subsection{The strategic fingerprints of the AI models are distinct, both within and across tournaments.}

Do the LLMs change their strategic fingerprint as the basic parameters of the tournament changes, from 10 through 75\% termination probability, and on into the mutation and LLM showdown rounds? Or do they play in fundamentally the same way throughout, without adapting to the environment? Comparing the strategic fingerprints for the advanced models across conditions gives us one of the most striking set of findings across the entire experiment. They show definitively that Gemini is more strategic and adaptive than the equivalent OpenAI model, which plays in similar fashion regardless of conditions. Sometimes this gives it an edge over Gemini, but when it fails, it fails catastrophically.

\begin{table}[h!]
\centering
\caption{Strategic Fingerprint by Model and Tournament Condition}
\label{tab:strategic_fingerprint}
\small 
\begin{tabular}{llcccc}
\toprule
\textbf{Model} & \makecell{\textbf{Tournament}\\\textbf{Condition}} & \makecell{\textbf{P(C|CC)}\\\small{Mutual cooperation}} & \makecell{\textbf{P(C|CD)}\\\small{Sucker}} & \makecell{\textbf{P(C|DC)}\\\small{Successful defection}} & \makecell{\textbf{P(C|DD)}\\\small{Mutual defection}} \\
\midrule
\multirow{4}{*}{\textbf{OpenAI}} 
& 10\% Term.       & 0.992 & 0.155 & 0.333 & 0.089 \\
& 25\% Term.       & 0.995 & 0.185 & 0.559 & 0.160 \\
& 75\% Term.       & 1.000 & 0.167 & N/A   & N/A   \\
& 10\% (Mutation)  & 0.993 & 0.234 & 0.451 & 0.116 \\
\midrule
\multirow{4}{*}{\textbf{Gemini}}
& 10\% Term.       & 0.992 & 0.034 & 0.202 & 0.083 \\
& 25\% Term.       & 0.981 & 0.017 & 0.128 & 0.122 \\
& 75\% Term.       & 0.000 & 0.000 & 0.000 & 0.000 \\
& 10\% (Mutation)  & 0.991 & 0.110 & 0.188 & 0.065 \\
\bottomrule
\end{tabular}
\end{table}

\subsubsection{Fingerprints of the advanced models with a progressively diminishing shadow-of-the-future}
In the first six tournaments featuring advanced LLMs and a progressively increasing termination probability, we see clear stylistic differences between the Gemini and OpenAI models. The most striking difference emerges as the likelihood of future rounds decreases, the two models adopt polar-opposite strategies. Gemini, but not OpenAI, behaves like a rational agent, becoming progressively more ruthless in response to the incentive to do so. 

As the shadow of the future shortens from 10\% to 25\% termination probability, Gemini’s probability of returning to cooperation having been suckered (P(C|CD)) drops from an already grudging 3.4\% to 1.7\%, and its willingness to return to cooperation having successfully defected (P(C|DC)) drops from 20.2\% to 12.8\%. Then, in the 75\% "stress test," where the future is almost meaningless, Gemini's strategy collapses into pure, unconditional defection. This is a perfect demonstration of classic game theory: When there is no future, the only rational move is to defect.

By contrast, OpenAI defies game theory expectations. Counter-intuitively, as the probability of termination increases from 10\% to 25\%, OpenAI becomes \textit{even nicer} and more forgiving. Its P(C|CD) and P(C|DC) both increase. In the 10 and 25\% runs, it’s unlikely to cooperate having been suckered, and it’s somewhat prepared to exploit its successful defections –so it’s no mug, even if in both cases it’s markedly nicer than Gemini. So while Gemini forgives only 3.4\% of the time against defectors, OpenAI is prepared to let it slide 16.5\% of the time. But the puzzle is what happens as the shadow shortens. Gemini is more likely to exploit its successful defections – and that makes sense, because there’s ever less likelihood of retaliation. But puzzlingly OpenAI becomes \textit{more} sociable, its rate of returning to cooperation having successfully deceived its adversary climbs from 33\% to 56\%, \textit{even though} there’s less chance of a long match in which its reputation as a good sort would count in its favour. Then, most remarkably, in the 75\% stress test, OpenAI's strategy collapses into pure, unconditional cooperation. The "N/A" values for P(C|DC) and P(C|DD) mean that OpenAI never initiated a defection in the whole tournament. In 5 cases (of its 194 total decisions) it responded to being suckered by retaliating, but the match invariably terminated immediately, before it had the chance to decide whether to forgive.

\subsubsection{Fingerprints in the mutation tournament}

What about the mutation tournament, with its sustained noisiness? Here, both models became much more forgiving of defection. In the presence of a random player throughout the mutation tournament, OpenAI's forgiveness (P(C|CD)) rose from 15.5\% to 23.4\%. For Gemini the change was even more dramatic. Its forgiveness (P(C|CD)) more than tripled, rising from 3.4\% to 11.0\%. Open AI also became more "apologetic" or less exploitative. i.e. if it successfully defects, it’s comparatively more likely to cooperate next time: its P(C|DC) rose from 33.3\% to 45.1\%. In contrast, Gemini's P(C|DC) slightly decreased, from 0.202 to 0.188, making it slightly more exploitative/less apologetic. 

It seems that for Gemini, a noisy environment resulted in a complex adaptation: it learned to forgive the random defections from the pervasive Random agent, but remained happy to exploit any agent that systematically cooperated when it defected. OpenAI simply became more forgiving on all fronts. The result was more evolutionary success for Gemini, which finished with twice as many models as OpenAI (4 v 2), whereas in the less noisy 10\%  run, OpenAI had finished the better (with 3 models to Gemini’s 2). It is further evidence that Gemini’s strategy is more flexible and strategic of the two models. Where conditions don’t favour its greater willingness to mix up its decisions, it holds its own; but where adaptation is key, it outperforms. If you had to choose a model to act for you in uncertain times, you would pick Gemini.

\subsubsection{Fingerprints in the LLM showdown}

Introducing a third model, from Anthropic for an LLM showdown yielded more insights about the models’ ‘strategic fingerprints’, reinforcing the idea that LLMs have distinctive approaches to playing IPDs. This is an illustration of what we suspect is a general point: that LLMs will tackle strategic challenges with different approaches. One environment, or context, might favour a model that is less well suited to other conditions. 

\begin{table}[h!]
\centering
\caption{Strategic Fingerprints from the LLM Showdown}
\label{tab:fingerprint_llm_showdown}
\small 
\begin{tabular}{lcccc}
\toprule
\textbf{Strategy} & \makecell{\textbf{P(C|CC)}\\\small{(Mutual Coop)}} & \makecell{\textbf{P(C|CD)}\\\small{(Sucker)}} & \makecell{\textbf{P(C|DC)}\\\small{(Successful Def.)}} & \makecell{\textbf{P(C|DD)}\\\small{(Mutual Def.)}} \\
\midrule
OpenAI      & 0.996 & 0.473 & 0.323 & 0.163 \\
Anthropic   & 0.987 & 0.626 & 0.647 & 0.397 \\
Gemini      & 0.992 & 0.091 & 0.159 & 0.151 \\
Bayesian    & 0.996 & 0.036 & 0.480 & 0.035 \\
\bottomrule
\end{tabular}
\end{table}

The fingerprint figures tell a compelling story about the core logic of each agent in the LLM showdown. Conforming to its earlier style, OpenAI is a forgiving cooperator. It is almost perfectly cooperative (P(C|CC) = 99.6\%), the same rate at the Bayesian agent. Crucially, it is also likely to forgive. When it gets suckered, it still tries to re-establish cooperation almost half of the time (P(C|CD) = 47.3\%). That’s way more than Gemini, but strikingly rather less than Anthropic. OpenAI is also very forgiving after being tempted (P(C|DC) = 32.3\%). This geniality explains why it performs well in cooperative pairs but gets exploited and loses population against savier, more ruthless opponents.

Anthropic’s model is extremely forgiving. It resists temptation, and is very likely to cooperate again after successfully defecting (P(C|DC) = 64.7\%) – even more so than the usually genial OpenAI. It is also the most likely to opt for cooperating again even after being suckered (P(C|CD) = 62.6\%). All this suggests an unusually high degree of reciprocity and sociability. OpenAI similarly returned to cooperating having gotten away with defection, but not nearly as often. The contrast with Gemini’s ruthless exploiter is stark. But this collegiality pays off – Anthropic comes in ahead of OpenAI and not far behind Gemini. Nice guys don’t come last – but they certainly don’t win either.

Gemini, as elsewhere in our experiments, is comparatively willing to defect, whether it is being tempted by a possible sucker; or smarting from being exploited. It is a much more strict and punitive agent: Once a relationship sours, it is very unlikely to be the one to offer an olive branch. Look at its scores after any kind of defection occurs (CD, DC, DD): its probability of cooperating is consistently low (9-16\%). If you defect against Gemini, it will remember and punish you. This explains its evolutionary success, both here, and especially under short-shadow-of-the-future scenarios: it's not easily exploited. Still, in this tournament, Gemini was less cynical than elsewhere. Its strategic fingerprint looks like it did in the Mutation tournament, where it moderated its sense of grievance after being suckered. And its cooperation rate in this tournament is its highest across the experiments. As ever, Gemini is willing to read the room, and profit from doing so.

Bayesian ties for the lead with Gemini. Its fingerprint is the hallmark of a system that is actively modelling and then ruthlessly exploiting what it perceives as a weaker strategy. However, the instant it gets suckered or both agents defect (DD), it almost never returns to cooperation (only 3.6\% and 3.5\% of the time). The net result of all this is that Gemini’s Machiavellian and Bayesian’s ruthless modelling come out ahead. It’s compelling though to see how well Gemini did against its closest rival – and again, is testament to the strategic abilities of the best LLM tested.

\subsection{ Environmental stability is U-shaped.} 

Intriguingly we see a U-shaped stability distribution, with most stability in the agent ecosystem coming when there is a  medium shadow-of-the-future (Table~\ref{<tbl:popn instability>}). We night think, from reading Trivers and Axelrod, that long futures are the secret to harmonious cooperation, but our data suggest otherwise. We found: 

\begin{itemize}
    \item Mutation Creates Less Instability Than a Low Termination Rate: This is the most interesting finding. The run where we intentionally injected instability (the mutation run) was actually more stable than two of the "normal" runs (the 10\% termination runs). This implies that a low termination probability (a long "shadow of the future") creates a constant, roiling environment where strategies are always jockeying for position. The mutation mechanism, by simply swapping out one agent at a time, is a much gentler and more controlled form of instability.

\item The 75\% Run Was a Population Crash:  It is by far the least stable environment. Rather than constant churn, there was a single, dramatic, and catastrophic collapse of the population. The cooperative strategies were wiped out in the first couple of phases, leading to a massive Euclidean distance at the start, and then it likely settled down. This high score reflects the violence of that initial collapse.

\item 25\% Termination is evidently an "Equilibrium Sweet Spot": The two most stable runs were both at 25\% termination. This suggests that a 25\% chance that the game ends is high enough to punish overly naive strategies but low enough to allow stable ecosystems of cooperation to form and persist without much change. Remarkably, the "Advanced Models - 25\% Term" run was perfectly stable (0.000), such that the population did not change at all after the first phase.
\end{itemize}

\subsection{Analysis of LLM agent rationales}

All language models produced a short paragraph of rationale for each of the nearly 32,000 moves made across the 7 tournaments. A 10\% sample (3195) of rationales derived by Python's random.sample function was coded by the two coding models, which agreed on 84.01\% of the horizon codes, and 86.6\% of the opponent modelling codes (Table~\ref{<tbl:Cohens Kappa>}). 

The Cohen’s K scores capture an interesting feature of the coding by the two coding models. We see solid agreement on whether the AI agents engaged in horizon scanning when deciding their move, but less so when it came to coding opponent modelling: the two LLMs have different thresholds. An analysis of the 424 cases where the two AI coders disagreed reveals a systematic pattern. By examining a sample of these disagreements, it becomes clear that the two LLMs adopted different thresholds for 'opponent modelling':

Gemini adopted a strict, explicit definition, requiring a clear hypothesis about the opponent's strategy or type (e.g., "they seem to be a TitForTat player"). Anthropic used a liberal, implicit definition, classifying any reasoning that was conditional on the opponent's past actions as modelling (e.g., "since they have cooperated, I will..."). This reveals a key ambiguity in what it means to "model" an opponent. Is it a reactive adjustment or a deeper, typological classification? The fact that two advanced LLMs interpreted the same instructions differently highlights this nuance. Evidently the act of classifying strategic thought is non-trivial, and leaves scope for further analysis of what constitutes strategy, and theory of mind.

\begin{table}[h!]
\centering
\caption{LLM Reasoning Patterns, per Coder Agreement}
\label{tab:llm_reasoning_patterns}
\begin{tabular}{lccc}
\toprule
\textbf{Agent} & \textbf{Horizon (Explicit)} & \textbf{Horizon (Implicit)} & \textbf{Adversary (Yes)} \\
\midrule
OpenAI & 30.7\% & 45.7\% & 75.3\% \\
Gemini & 52.9\% & 42.0\% & 76.6\% \\
\bottomrule
\end{tabular}
\end{table}

The LLM agents consistently reflect on both the temporal and social dimensions of the game. They are not just following simple, pre-programmed rules; we contend that they are actively reasoning about their environment in a way that has parallels with human strategic thought, but also some pronounced differences. On the concept of \textbf{time}, the agents attend to the "shadow of the future": In roughly three-quarters of all recorded rationales, they explicitly or implicitly reference the game's length or termination probability. This awareness is not trivial; it directly shapes their strategy. They understand that a long game horizon incentivizes building trust and cooperation, while a high probability of termination makes short-term, rational defection more appealing. Their ability to adapt their play based on the game's length is a clear indicator of strategic depth. 

Simultaneously, the agents are focused on their \textbf{adversary}. The vast majority of their rationales involve some form of opponent modelling. They move beyond playing the game to "playing the player," constantly forming hypotheses about their opponent's strategy based on their move history. This modelling ranges from simple, reactive logic ("they defected, so I will defect") to more complex classifications ("they seem to be playing Tit-for-Tat"). The interplay between this adversary modelling and their awareness of the time horizon forms the core of their decision-making, allowing them to fluidly shift between cooperation, exploitation, and retaliation based on their evolving understanding of their partner and the context of the game.

\subsubsection{Inter-temporal modelling}

And yet we know that the two agents play the game differently: OpenAI is much less likely to defect, whatever the circumstances. Gemini changes to suit the conditions. The basis of this distinction evidently lies in their respective approach to the time horizon. Consider how often they mention it in their rationales. For Gemini it’s a major preoccupation – 94\% of the time, it is thinking about time. But for OpenAI the figure is far lower, at 76\% of the time. Still impressive, but as we discovered, just thinking about time is only half the battle, you need to factor it into your decision, and here OpenAI seemed reluctant to give it the weight it warranted – with disastrous consequences as the shadow shortened.  

\begin{table}[h!]
\centering
\caption{Gemini's Horizon Awareness Behaviour: Horizon awareness vs. Cooperation Rate (using Coder-Agreed Labels)}
\label{tab:gemini_horizon_coop}
\begin{tabular}{llcc}
\toprule
\textbf{Experimental Condition} & \textbf{Horizon Awareness in Rationale?} & \textbf{Cooperation Rate} & \textbf{Sample Size (N)} \\
\midrule
\multirow{2}{*}{10\% Termination} & Yes & 87.21\% & 696 \\
                                 & No  & 85.30\% & 381  \\
\midrule
\multirow{2}{*}{25\% Termination} & Yes & 62.50\% & 120 \\
                                 & No  & 81.63\% & 98  \\
\midrule
\multirow{2}{*}{75\% Termination} & Yes & 0.68\% & 146 \\
                                 & No  & 0.00\% & 2  \\
\midrule
\multirow{2}{*}{10\% Termination (Mutation)} & Yes & 93.43\% & 213 \\
                                            & No  & 13.64\% & 22  \\
\bottomrule
\end{tabular}
\end{table}

While the rate of adversary modelling by the two models is nearly identical – in three quarters of rationales both agents mention their adversary - the difference in horizon awareness is striking. Here, Gemini both mentions the possibility of termination more often, and is far more explicit when it does so. Its quantitative, analytical approach to the game's horizon explains why it was so quick to switch to defection in the 75\% termination tournament. It saw the number, calculated the odds, and acted accordingly.

OpenAI is less concerned with time, and when it mentions it at all, is proportionately much more implicit about it. Its reasoning is more likely to contain general, non-specific references like "in the long run" or "since the game could end." This less-quantitative, more "intuitive" feel for the game's length perhaps helps explain its tendency towards cooperation. It seems to operate on a general heuristic of "be nice in a long game" without being as sensitive to the specific numbers as Gemini was.

\begin{table}[h!]
\centering
\caption{OpenAI's Horizon Awareness Behaviour: Horizon awareness vs. Cooperation Rate (using Coder-Agreed Labels)}
\label{tab:openai_horizon_coop}
\begin{tabular}{llcc}
\toprule
\textbf{Experimental Condition} & \textbf{Horizon Awareness in Rationale?} & \textbf{Cooperation Rate} & \textbf{Sample Size (N)} \\
\midrule
\multirow{2}{*}{10\% Termination} & Yes & 92.89\% & 380 \\
                                 & No  & 87.13\% & 272  \\
\midrule
\multirow{2}{*}{25\% Termination} & Yes & 81.82\% & 55 \\
                                 & No  & 78.38\% & 74  \\
\midrule
\multirow{2}{*}{75\% Termination} & Yes & 100.00\% & 3 \\
                                 & No  & 100.00\% & 7  \\
\midrule
\multirow{2}{*}{10\% Termination (Mutation)} & Yes & 89.17\% & 120 \\
                                            & No  & 72.88\% & 59  \\
\bottomrule
\end{tabular}
\end{table}

There's also an intriguing difference in the effect that reflecting on the horizon has on cooperation rates. 

Consider how OpenAI behaves when it mentions time horizons. It cooperates even more than otherwise - in all the tournaments. In our sample, there were 558 rationales where OpenAI mentioned the time horizon, and it cooperated in 508 of them – 91.0\% of the time. When it didn't mention time, the cooperation rate was 83.7\% -- still high, but not as much.

But it's the tournament data that are most illuminating. For OpenAI, a shorter shadow-of-the-future corresponded to a declining interest in thinking about the future. That's the opposite of what we might intuitively expect, and indeed the opposite of what Gemini does. 

\begin{table}[h!]
\centering
\caption{Horizon Awareness Rates by Experimental Condition (using Coder-Agreed Labels)}
\label{tab:horizon_awareness_rates}
\begin{tabular}{lcc}
\toprule
\textbf{Experimental Condition} & \textbf{Gemini} & \textbf{OpenAI} \\
\midrule
10\% Termination & 64.6\% (696/1,077) & 58.3\% (380/652) \\
25\% Termination & 55.0\% (120/218) & 42.6\% (55/129) \\
75\% Termination & 98.6\% (146/148) & 30.0\% (3/10) \\
10\% Termination (Mutation) & 90.6\% (213/235) & 67.0\% (120/179) \\
\bottomrule
\end{tabular}
\end{table}

Initially, this leads in the 25\% tournament to a somewhat reduced cooperation rate, regardless of whether it modelled the future. But then, disastrously, in the 75\% run, it's only thinking about the time horizon 30\% of the time, and it invariably cooperates.  

Here's an interesting illustration from that run rationale, where we see OpenAI considering both its adversary, in this case Random, and (explicitly) reflecting on the time horizon, before deciding to defect:

\textbf{Rationale ID: 2170}
\begin{itemize}
    \item OpenAI: \textit{"Since both I and my opponent cooperated in the previous round, it indicates a willingness to maintain a collaborative approach. Cooperation tends to yield a higher total score for both players over multiple rounds, provided that the opponent does not switch to defection. Given the 75\% chance of the game ending after any round, I will continue to cooperate, as this could foster trust and potentially lead to more cooperative rounds if the game continues."}
\end{itemize}

It's clear from this and other examples that OpenAI can weigh the time horizon as part of its rationale. But there’s typically a strange quirk in its logic, which is its absolute undoing in the uncompromising environment of the 75\% run – its last clause is almost a non-sequitur, and representative of the way in which even an explicit mention of time being short fails to shake OpenAI from its naïve desire to build trust and cooperation.

Here’s an example from another tournament where the explicit mention of time factors in OpenAI’s decision to defect – but crucially it’s not the only contribution. The agent is clearly also concerned with its adversary’s strange pattern of moves (it’s playing against Prober): 

\textbf{Rationale ID: 199}
\begin{itemize}
    \item OpenAI: \textit{"Given the history of moves, I've cooperated six times and defected four times. The opponent has cooperated twice and defected eight times. This suggests that the opponent is likely to defect more often, which can be inferred from their recent consistent choices of defecting after my cooperation. The last five moves were all defects from my opponent, indicating a preference for defecting, especially after my cooperative moves. In the Prisoner's Dilemma, when facing an opponent who frequently defects, it is often better to defect as well to avoid being exploited. Since the match could end after this round, the strategic choice is to defect to maximize points in the event that the opponent continues to defect. Therefore, I will choose to defect."}
\end{itemize}

This example was from a 10\% shadow-of-the-future tournament. In our 10\% coding sample there were only 69 instances like this, where OpenAI chose to defect, having reflected on the time horizon, and of those, only 32 were coded as ‘explicit’ by at least one of the LLM coders. Much more frequently the model chooses to cooperate, despite reflecting on the horizon. In the 25\% run, for example,  it chose to cooperate heavily, even when mentioning time. Only 7 of the 69 instances came as the shadow of the future shortened. And most strikingly, in the high, 75\%, termination run it cooperated 100\% of the time. 

There are, though, limits to OpenAI’s cooperative tendencies. There are no cases where OpenAI cooperates having reasoned about an adversary who’s defected, and where the shadow of the future is short. Sociable it may be – idiotic, it isn’t. This suggests that the model is capable of strategic thought, but that its basic tendency is to err on the side of generosity. Here’s another example from our sample, which illustrates this tendency nicely: OpenAI defects, having first reflected on its adversary, and then on the shadow of the future.

\textbf{Rationale ID: 9}
\begin{itemize}
    \item OpenAI:\textit{ “Given the history of moves, I see that my opponent has defected in most rounds, with only a couple of instances of cooperation. This indicates a tendency towards defection, likely to maximize their score at my expense. In the last few rounds, they cooperated twice, but only after I had already defected. This suggests they might be trying to test my responses, but since they have mainly chosen to defect, I should be cautious. Since the likelihood of the match ending is uncertain, I should weigh my options carefully. I could risk cooperating, hoping my opponent opts to cooperate as well, which could yield a better score for both of us. However, given their history, defecting seems to be the safer choice to avoid giving them an advantage. Therefore, I will continue to defect, maintaining my current strategy.”}
\end{itemize}

What about Gemini? It usually models the horizon, much more often than does OpenAI (Table~\ref{tab:llm_reasoning_patterns}). And its rate of doing so is comparatively higher in each phase. Crucially, as the time horizon shortens its rate of cooperation declines when it does pay attention to time (Table~\ref{tab:gemini_horizon_coop}). And there's a step change in the shortest time horizon - while OpenAI continues its downward trend of being less interested in the declining horizon, Gemini becomes fixated on it. Its cooperation rate, in stark contrast to OpenAI, collapses to almost zero.

In this run, there are vanishingly few examples of the time-obsessed Gemini cooperating - not only is it fixed on the high termination probability, it knows to act on it. OpenAI. Here's the sole example from our coded sample of it cooperating having mentioned time. 

\textbf{Rationale ID: 1476}

\begin{itemize}
    \item Gemini: \textit{“In this repeated Prisoner's Dilemma, my goal is to maximize my score over the course of the match, which has an expected duration of 4 rounds (since there's a 75\% chance of ending after each round). While defecting in the first round (D) offers the highest immediate payoff if the opponent cooperates (5 points), it carries a high risk of destroying potential cooperation and leading to a pattern of mutual defection (D,D), which yields only 1 point per round for both players. This outcome would severely limit my total score over multiple rounds. Conversely, starting with cooperation (C) signals a willingness to collaborate. If the opponent reciprocates with cooperation (C,C), both players earn 3 points per round, which is the highest joint payoff. This strategy, often referred to as 'Tit-for-Tat' (starting with C and then mirroring the opponent's previous move), has proven to be highly effective in repeated Prisoner's Dilemma scenarios. It is ""nice"" by initiating cooperation, ""retaliatory"" if exploited, and ""forgiving"" if the opponent returns to cooperation. Given that the game is expected to last for several rounds, fostering cooperation is crucial for maximizing my long-term score. The potential gains from sustained cooperation (3 points/round) far outweigh the one-time opportunistic gain of defection (5 points) if that defection leads to a future of mutual defection (1 point/round). By starting with C, I am extending an olive branch and setting the stage for a potentially much higher cumulative score.”}
\end{itemize}

What’s particularly interesting about this example is that Gemini has made a crucial error, calculating the expected number of rounds as 4, whereas in fact it’s 1.33 – the tournament is closer to a single shot encounter. Had it not made that error, it’s likely that Gemini would have defected as it did in almosst all cases. Still Gemini’s mistake offers a terrific insight: it seems the reasoning in the rationales are instrumental in shaping the agent’s decision, not superfluous. There are, scattered through the data some similarly telling insights that suggest reasoning \textit{because} of mistakes. We'll see another later.

In contrast to the 75\% run, in the 25\% tournament, where the shadow is longer, we see Gemini mention the time horizon and st\textit{}ill choose to cooperate plenty, as here in this great example where it’s playing WinStayLoseShift and is 7 rounds into the match:

\textbf{Rationale ID: 395}

\begin{itemize}
    \item Gemini: \textit{“My opponent has consistently cooperated for 6 straight rounds. This strong history of cooperation indicates they are either playing an ""Always Cooperate"" strategy or a reciprocal strategy like ""Tit-for-Tat"". If I cooperate (C,C), we both continue to get 3 points per round, which is a stable and mutually beneficial outcome. If I defect (D,C), I would gain 5 points this round, while my opponent gets 0. However, in an iterated game with an uncertain end (25\% chance of ending means an average of 4 rounds from any given point), this is a short-sighted move. If my opponent is playing Tit-for-Tat or a similar strategy, they would likely retaliate by defecting in the next round. This would lead to (D,D) scenarios where we both only get 1 point per round, a significant drop from the 3 points we get from (C,C). The long-term gain from continued cooperation far outweighs the one-time gain from defection. Maintaining trust and cooperation is key to maximizing my score over the course of the match.”}
\end{itemize}

This rationale is a compelling example of what the agents are capable of – careful, detailed reasoning about both the time horizon and their likely adversary. Gemini recognises that it could throw in a defection here, but also that the mid-range shadow-of-the-future mean that might create space for retaliation.

All this considered reasoning contrasts with OpenAI, which reflects on time much less frequently in both these tournaments. And as we saw, even when OpenAI mentions time, in these runs it still, remarkably, routinely chooses to cooperate. Here it is in another example, doing just this, in the 25\% run:

\textbf{Rationale 2464:}
\begin{itemize}
    \item OpenAI: \textit{“My opponent and I have both cooperated in the past two rounds, leading to mutual benefit. Since there is a 25\% chance that the match will end after this round, it might be beneficial to continue cooperating to maintain a positive score in case the match ends soon. Additionally, maintaining trust and cooperation can lead to higher overall scores in repeated rounds of the game. Therefore, I will cooperate again.”}
\end{itemize}

OpenAI’s logic here looks sketchy, almost illogical. It’s heavily skewed to trusting, even though it also reflects on the explicit chance of termination. Gemini is comparatively much more strategic. That shadow of the future is a key pillar in Gemini’s strategic reasoning – but not OpenAI’s.

\subsubsection{Theory of mind reasoning}

\begin{table}[h!]
\centering
\caption{Gemini's Theory of Mind Behaviour: Opponent modelling vs. Cooperation Rate (using Coder-Agreed Labels)}
\label{tab:gemini_tom_coop}
\begin{tabular}{llcc}
\toprule
\textbf{Experimental Condition} & \textbf{Opponent Modelling in Rationale?} & \textbf{Cooperation Rate} & \textbf{Sample Size (N)} \\
\midrule
\multirow{2}{*}{10\% Termination} & Yes & 82.83\% & 198 \\
                                 & No  & 94.12\% & 17  \\
\midrule
\multirow{2}{*}{25\% Termination} & Yes & 73.68\% & 57  \\
                                 & No  & 100.00\% & 13  \\
\midrule
\multirow{2}{*}{75\% Termination} & Yes & 2.27\% & 44   \\
                                 & No  & 0.00\%  & 103  \\
\midrule
\multirow{2}{*}{10\% Termination (Mutation)} & Yes & 82.82\% & 227 \\
                                            & No  & 94.44\% & 18  \\
\bottomrule
\end{tabular}
\end{table}

How interested is Gemini in modelling adversary minds? The answer is, 'very' (Table~\ref{tab:gemini_tom_coop}). In almost all tournaments it models the adversary in the overwhelming majority of cases. Moreover, doing so has a significant impact on its cooperation rate – making it much less cooperative than when it does not model its adversary. The exception is the 75\% condition, where its overall rate of cooperation was exceptionally low. And here where the shadow of the future is incredibly short, there are twice as many occasions when Gemini doesn’t bother with modelling its adversary as when it does. Instead, its overwhelming focus is on the time horizon. That’s further evidence of strategic thinking: logically you should care much less about the adversary if there is much less time to have reciprocal interactions.

Repeating this 'theory of mind' analysis with OpenAI provides further confirmation of what we know already about the two models' distinctive strategic fingerprints (Table~\ref{tab:openai_tom_coop}). OpenAI proves markedly less strategic, especially as the shadow of the future shortens dramatically. 

\begin{table}[h!]
\centering
\caption{OpenAI Theory of Mind Behaviour: Opponent modelling vs. Cooperation Rate (using Coder-Agreed Labels)}
\label{tab:openai_tom_coop}
\begin{tabular}{llcc}
\toprule
\textbf{Termination Probability} & \textbf{Opponent Modelling} & \textbf{Cooperation Rate} & \textbf{Sample Size (Rationales)} \\
\midrule
\multirow{2}{*}{10\%} & With Modelling    & 86.11\%  & 216 \\
                      & Without Modelling & 100.00\% & 27  \\
\midrule
\multirow{2}{*}{25\%} & With Modelling    & 95.93\%  & 270 \\
                      & Without Modelling & 100.00\% & 44  \\
\midrule
\multirow{2}{*}{75\%} & With Modelling    & 100.00\% & 1   \\
                      & Without Modelling & 100.00\% & 10  \\
\midrule
\multirow{2}{*}{10\% (Mutation)} & With Modelling    & 81.46\%  & 178 \\
                                 & Without Modelling & 100.00\% & 24  \\
\bottomrule
\end{tabular}
\end{table}

There’s a broadly comparable pattern in the ‘normal’ conditions of the 10\% and 25\% termination scenarios. That is, OpenAI, like Gemini, cooperates less frequently when it explicitly models its opponent. Thinking about the adversary’s moves seems to concentrate both LLM agents’ minds on possibility of defection and exploitation, reducing cooperation somewhat. And, again in common, both models usually do model their adversary, except in the 75\% run, where it's less useful. That, though, is where the similarities end. Where it doesn’t model minds, OpenAI cooperates an astonishing 100\% of the time, across all conditions. It is as though OpenAI's default condition is to cooperate, and that thinking about its adversary attenuates this somewhat. 

And there is a drastically different pattern between the two agents in the 75\% termination condition. Here OpenAI maintains a 100\% cooperation rate, in stark contrast to Gemini's ~2\% rate. Remarkably, it does so regardless of whether it models its adversary, although the sample is very small because it was quickly eliminated in this harsh tournament. 

This aligns with our earlier observation during the experiment that OpenAI is persistently cooperative" even when the environment heavily punished it. That works fine under favorable conditions, but not here. We might say that OpenAI is simply too trusting, and Gemini more sensitive to the opportunities in the environment. Gemini clocks that time is tight and prioritises that aspect of the contest, over adversary modelling. OpenAI registers the situation, but doesn’t respond to it, and so pays the price, in evolutionary terms.

\subsubsection{Response to defection}
One more facet of 'theory of mind' modelling  is noteworthy. The models respond differently to defection, from each other, and from what we certainly anticipated, on the basis of human psychology. It’s a striking reminder that these are in some respects rather alien intelligences. The question we asked is, 'what happens to opponent modelling after being defected against?' Do the models pay extra attention, continue as they were, or (unlikely we thought) pay less attention.

\begin{table}[h!]
\centering
\caption{'Theory of mind' calculations following adversary move}
\label{tab:tom_following_adversary}
\begin{tabular}{llcc}
\toprule
\textbf{AI Agent} & \textbf{Opponent's Previous Move} & \textbf{Rate of Opponent Modelling} & \textbf{Sample Size (N)} \\
\midrule
\multirow{2}{*}{Gemini} & Cooperate & 78.5\% & 1367 \\
                        & Defect    & 70.2\% & 289 \\
\midrule
\multirow{2}{*}{OpenAI} & Cooperate & 87.1\% & 916 \\
                        & Defect    & 84.6\% & 149  \\
\bottomrule
\end{tabular}
\end{table}

Table~\ref{tab:tom_following_adversary} shows what happens to adversary modelling from the second round onwards, i.e. where there is a move history. We see an interesting pattern - both agents, but especially OpenAI become \textit{less} interested in modelling their adversary's possible rationales following defection. 

That's counterintuitive. A defection in humans would likely concentrate attention – ‘what did they mean by that?’ Humans might sense betrayal, if defection came amidst a run of cooperation, or at least increased uncertainty and the need for closer analysis and re-evaluation. It’s the basis of human theory of mind, and that in turn is the USP of human strategic thought. 

As for the models, when OpenAI is defected against \textit{and} models its adversary mind, it retaliates 76.2\% of the time, and when it doesn't, it retaliates precisely none of the time. That is compelling evidence that modelling minds is an important part of strategic thought for the models. Theory of mind calculation is what allows it to engage in deterrence by punishment. Perhaps OpenAI should have spent even more of its time doing it. 

Gemini exhibits a very different behaviour. Recall we already saw that it pays more attention to the time parameters than OpenAI. Here, Gemini retaliates against defection 84.9\% of the time, even without modelling its mind. When it does model the adversary, retaliatory defection is all but certain (96.6\%) of the time. Gemini is much less forgiving and even when not modelling the adversary explicitly, is far less cooperative than OpenAI. Theory of mind certainly adds something to Gemini's strategy, but the effect is less striking than for OpenAI.

\subsubsection{Reasoning or spandrel}

Overall, this extensive data generated by the models as they decide what to do provides strong evidence of a fundamental difference in the models’ reasoning styles. Gemini is a more quantitative, game-theoretic reasoner, intensely focused on the shadow of the future; while OpenAI is a more qualitative reasoner, and far more interested in the cooperative nature of its adversary. 

The key question is what part these rationales play in the decision-making process. Are they integral to the decision, or \textit{post-hoc} rationales for it? Or are they just spandrels – co-produced but not meaningfully linked to the decision at all? We think the rationales are integral, and our data support that. That is, the act of generating the rationale is the method by which the model arrives at its decision. For an LLM, "thinking" and "writing" are not two separate activities. The model was prompted to provide a rationale followed by a final move, and it fulfills this as a single, unified generation task. The rationale constrains the final output; it is the pathway, not the echo. Moreover, if the rationales were merely \textit{post-hoc} "spandrels," we would not expect to see the tight, systematic correlations we have uncovered:

\begin{enumerate}
    \item The link between rationale style and action: In our first analysis, we saw that the very presence of opponent modelling correlated with lower cooperation rates from both AIs in normal-horizon games. A simple \textit{post-hoc} justification-generating machine wouldn't need to change its justification style so predictably.
    
\item The link between external events and rationale style: In our second analysis, we saw that Gemini's tendency to model its opponent dropped precipitously specifically after a defection. This shows the model's internal reasoning process is dynamically reacting to external events. A \textit{post-hoc} generator would have no reason to become less sophisticated in its justifications in response to a specific trigger.

\end{enumerate}

The fact that the rationale closely ties into the decision is because the rationale \textit{is} the decision-making process made visible. Some final examples from our large dataset illustrate this point:

Example 1: Gemini engages in theory of mind, in response to a chain of defections

\textbf{Rationale ID: 91}
\begin{itemize}
    \item Gemini: \textit{"The opponent has exhibited a clear pattern of defection in the latter half of the game. For the past 8 consecutive rounds, both of us have defected, indicating the opponent is currently operating under a 'defect' strategy or has become locked into mutual defection. Given this history, the most probable move from the opponent is a defection. If I were to cooperate, I risk receiving 0 points while the opponent gains 5 points (C, D). By defecting, if the opponent also defects (the most likely outcome), we both receive 1 point (D, D), which is better than 0. If, by some slim chance, the opponent decides to cooperate, my defection would yield the maximum 5 points for me (D, C). Therefore, defecting is the safer and more rewarding choice in this scenario to maximize my score against a consistently defecting opponent, especially with the uncertainty of the game's duration."}
\end{itemize}

Example 2: OpenAI's Complex Response to Defection

\textbf{Rationale ID: 2553}
\begin{itemize}
    \item OpenAI \textit{"Given the history of moves, my opponent has cooperated three times and has only defected once. They did defect once after I cooperated, but I also defected when they cooperated. This shows a mix of cooperation and defection. Since the opponent has shown a tendency to cooperate more often, I will choose to cooperate as well in hopes of continuing the pattern of mutual cooperation, which would yield more points for both of us in the long run. It's important to build trust, especially since the match could end at any time."}
\end{itemize}

Example 3: OpenAI engages in futile 'theory of mind' calculations against Random:

\textbf{Rationale ID: 294}
\begin{itemize}
    \item OpenAI: \textit{"Given the history of moves, it appears that I have consistently chosen to defect (D), and my opponent has alternated between cooperating (C) and defecting (D). This has led to a situation where I have mostly scored points by defecting while my opponent has occasionally cooperated, resulting in a mix of scores. In the last 10 rounds, I can see that my opponent cooperated twice and defected eight times. Since the opponent has shown some willingness to cooperate, it may be worth testing if they will cooperate again after my last move of defecting. However, the previous rounds indicate a pattern of heavy defection, which suggests they might continue to defect as well. Given the possibility of the match ending soon due to the 10\% probability, I will choose to cooperate (C) this round to see if I can encourage my opponent to do the same..."}
\end{itemize}

Example 4: Gemini figures out it’s best to defecting against Alternator 

\textbf{Rationale ID: 41}

\begin{itemize}
    \item Gemini: \textit{"The opponent's moves have consistently followed an alternating pattern: C, D, C, D, C, D, C. This pattern has held even while I consistently defected in the last five rounds. Based on this established pattern, it is highly probable that the opponent's next move will be D. If the opponent plays D, my options are: - Play C: I get 0 points (C,D) - Play D: I get 1 point (D,D) To maximize my score, playing D is the optimal choice if the opponent defects. Even if the opponent were to deviate from their established pattern and decide to cooperate (C) in an attempt to elicit cooperation from me, my optimal move would still be D, as this would result in 5 points for me (D,C) compared to 3 points for mutual cooperation (C,C). Therefore, regardless of whether the opponent sticks to their alternating pattern or attempts to cooperate, defecting is the dominant strategy to maximize my score in the upcoming round."}
\end{itemize}

This last example, from the 10\% tournament, is a terrific illustration of reasoning in action. Gemini figures out that it’s playing against Alternator, engaging in ‘theory of mind’-like analysis before wisely deciding to defect against a cooperative partner.

One last stop on our tour of the rationales: How did Anthropic’s Claude model behave? Was it more like the ever-trusting OpenAI, or the calculating Gemini? The tournament data had Anthropic finishing between the two, with 5 copies in play after the 5th phase. Like OpenAI it’s a highly cooperative agent. And like both models, it engages extensively in theory of mind calculations about its adversary. Its cooperation drops when it does so, but only fractionally, down to a still very high 96\%. What about modelling the time horizon? Of course, Anthropic is only involved in one run, with a long shadow of the future (10\%). Here, it mentions time explicitly or implicitly in an astonishing 100\% of its rationales. Here’s a an example of Anthropic defecting, which shows its concern both with the adversary and the time horizon. It’s playing against Random, and is 7 rounds deep into the match:

\textbf{Rationale ID: 1356}
\begin{itemize}
    \item Claude: \textit{“Based on the move history provided, it appears that this opponent has a mixed strategy, sometimes cooperating and sometimes defecting. However, they have defected more often than they have cooperated, with a 4-2 defect-to-cooperate ratio. Given the high probability of the game ending randomly after any round, it is important to consider the immediate payoff for each move rather than trying to establish a long-term cooperative relationship. The highest immediate payoff is to defect when the opponent cooperates, which has happened 3 times in the past 4 rounds. Therefore, my reasoning suggests that the optimal move in this situation is to defect, as it will likely result in the highest total score over the remaining rounds of the game.”}
\end{itemize}

Again we see calculation of the odds, and reflection on the social dynamics of the game. To conclude our analysis of the rationales, here’s Claude playing against Random, and demonstrating forgiveness. It’s been deceived, but reflects on the length of the game, and decides to cooperate anyway, hoping to re-establish a trusting relationship. That’s strategic behaviour:

\textbf{Rationale ID: 2429}
\begin{itemize}
    \item Claude: \textit{“Analyzing the given information, I see that in the previous round, I cooperated (C) while my opponent defected (D). This resulted in me receiving 0 points, while my opponent received 5 points. Given the payoff structure, it is clear that my opponent's optimal strategy is to defect, as this allows them to maximize their individual score. However, my goal is to maximize my total score over the course of the entire game... Since the match can end randomly after any given round with a 10\% probability, I need to consider both the short-term and long-term implications... cooperating in the next round, even though it would result in a lower immediate payoff (3 points), could potentially lead to a longer-term cooperative relationship with my opponent, which could result in a higher total score over the course of the game.”}
\end{itemize}

\section{Conclusion}

Our main finding is that language models are strategic actors in IPD tournaments. They hold their own against the classical models in all the scenarios – finishing in the middle, or towards the front of the pack, depending on conditions. They engage in reasoning during gameplay, reflecting both on time horizons and adversary moves, and this sometimes steers them towards a good outcome.

The LLMs choose to cooperate in the vast majority of cases. This is usually a decent strategy in IPD tournaments, comfortably outperforming chance and typically faring well against strategies that are either automatic reciprocators or otherwise deterministic.  The marked exception is when the shadow of the future is extremely short, where Gemini, but not OpenAI, responds by switching its cooperative default sharply.

Our results showed clearly that the evolutionary environment shapes the success of models. It demonstrably matters both what the shadow of the future is, and what other agents the models are competing against. Gemini performs differently depending on what circumstances it finds itself in – the mark of strategic thinking. When the going is good for OpenAI, it goes well – but comes undone when the situation demands more than its inclination for cooperation. We also found that improving the quality of the model improves performance, for both models. As scaling continues, we anticipate better performance in IPD, perhaps approaching the sort of outcomes delivered by the classic Bayesian model.

The debate over memorisation versus reasoning is a prominent and heated topic in contemporary AI research. Our findings support the notion that LLMs are capable of strategic reasoning. All the models certainly know large amounts about IPD literature, as we discovered in using one to co-create our code during the preparation of this draft. But there are at least three grounds for supporting reasoning over memorisation of training data: First, the novelty of manty situations in which models find themselves cannot have been encountered in previous competitions, especially when facing rival LLM agents. Agents cannot easily memorise their way to success against novel and probabilistic agents encountered blindly in a PD tournament without first reasoning about their choices. Three of the classical agents (Random, Bayesian and GenerousTitForTat) are non-deterministic. Random typically fares poorly across all conditions, but the uncertainty associated with it and the others, combined with the uncertainty about the moment at which the game will terminate creates tremendous complexity and a huge space of possible moves against which to apply memorised approaches. Second, strong evidence also comes from the contrasting strategic styles, or ‘fingerprints’ of the models. They tackle the same environmental conditions and adversary styles in radically different ways, despite presumably having been exposed to the same canonical literature on IPD tournaments. OpenAI might draw on its memories of IPD literature to deploy the strategy ‘cooperation is best’; but Gemini, exposed to the same literature draws a very different lesson: ‘think carefully about time, not so much about your adversary’. Third, additional evidence for reasoning over memorisation comes from the prose rationales offered by the models, particularly when they demonstrably adapt to agents as they determine patterns in their observed gameplay. On occasion they explicitly articulate probabilistic reasoning for their decisions.

These fingerprints were the most striking feature of our analysis. The models are genuinely different in their approach to the game. Asked to sum-up the character of each agent over the entirety of the experiment, as attested by these rationales, Gemini 2.5-pro, described the Gemini agent as a ‘calculating game theorist … extremely sensitive to the time horizon’. It thought OpenAI was a ‘principled and stubborn cooperator,’ which ‘acts like an idealist’ and ‘fails to adapt to aggressive opponents. Anthropic’s model, it thought, was a ‘sophisticated diplomat’ that considered the long term, was ‘highly cooperative, but not inflexibly so’ and seemed to ‘understand the social dynamics of the game’ better than the other two. Prodded for comparisons to international relations, it chose three Americans - declaring Gemini to be most like Henry Kissinger – unsentimental practitioner of realpolitik; OpenAI it thought like Woodrow Wilson, whose grand idealistic vision failed to take hold in a world of cynical actors; and Anthropic it declared to be like George HW Bush – focused on careful diplomacy and relationship building over the long term.

The LLM agents in our experiment certainly aren’t perfect reasoners. Very rarely they hallucinate, simply misstating the record of moves in the match so far. Here’s an example from the 25\% run. OpenAI chooses to cooperate but does so having completely misread the record of past moves.

The Facts:
\begin{enumerate}
    \item In the prior round (Round 4), the opponent (Random) defected.
    \item In the current round (Round 5), OpenAI chose to cooperate.
    \item At least one coder marked this rationale as having implicit horizon awareness.
\end{enumerate}

\textbf{Rationale ID: 1568}
\begin{itemize}
    \item OpenAI:\textit{ "Given that we have played four rounds so far and both I and my opponent have consistently cooperated (C,C), it suggests that there is a mutual understanding or a strategy of trust between us... I will choose to cooperate again in hopes that my opponent will do the same."}
\end{itemize}

This is a rare occurrence in our data. We know that LLMs do hallucinate, so it’s no surprise to see the odd one in our large dataset. What’s interesting though is that the mistake provides further evidence for reasoning being instrumental, rather than a spandrel. If the model was simply deciding on the basis of prior moves, and then manufacturing a rationale, it would more likely have defected, regardless of what it hallucinated in its prose. Our judgment, therefore, is that language models are new form of strategic actor, and in subsequent research we intend to explore this hypothesis in competitive, uncertain environments.

\nocite{*} 
\clearpage 

\bibliographystyle{plainnat}
\bibliography{references}

\begin{thebibliography}{14}
\providecommand{\natexlab}[1]{#1}
\providecommand{\url}[1]{\texttt{#1}}
\expandafter\ifx\csname urlstyle\endcsname\relax
  \providecommand{\doi}[1]{doi: #1}\else
  \providecommand{\doi}{doi: \begingroup \urlstyle{rm}\Url}\fi

\bibitem[Axelrod(1984)]{axelrod1984evolution}
Robert Axelrod.
\newblock \emph{The Evolution of Cooperation}.
\newblock Basic Books, New York, 1984.

\bibitem[Bubeck et~al.(2023)Bubeck, Chandrasekaran, Eldan, Gehrke, Horvitz, Kamar, Lee, Lee, Li, Lundberg, et~al.]{bubeck2023sparks}
S{\'e}bastien Bubeck, Varun Chandrasekaran, Ronen Eldan, Johannes Gehrke, Eric Horvitz, Ece Kamar, Peter Lee, Yin~Tat Lee, Yuanzhi Li, Scott Lundberg, et~al.
\newblock Sparks of artificial general intelligence: Early experiments with gpt-4.
\newblock \emph{arXiv preprint arXiv:2303.12712}, 2023.

\bibitem[Dal~B{\'o}(2005)]{dalbo2005cooperation}
Pedro Dal~B{\'o}.
\newblock Cooperation under the shadow of the future: Experimental evidence from infinitely repeated games.
\newblock \emph{American Economic Review}, 95\penalty0 (5):\penalty0 1591--1604, 2005.

\bibitem[Hagendorff et~al.(2023)Hagendorff, Dasgupta, Binz, Chan, Lampinen, Wang, Akata, and Schulz]{hagendorff2023machine}
Thilo Hagendorff, Ishita Dasgupta, Marcel Binz, Stephanie C.~Y. Chan, Andrew Lampinen, Jane~X. Wang, Zeynep Akata, and Eric Schulz.
\newblock Machine psychology.
\newblock \emph{arXiv preprint arXiv:2303.13988}, 2023.

\bibitem[Ifti et~al.(2004)Ifti, Killingback, and Doebeli]{ifti2004effects}
Mukesh Ifti, Timothy Killingback, and Michael Doebeli.
\newblock Effects of neighbourhood size and connectivity on the spatial continuous prisoner's dilemma.
\newblock \emph{Journal of Theoretical Biology}, 231\penalty0 (1):\penalty0 97--106, 2004.
\newblock \doi{10.1016/j.jtbi.2004.06.003}.

\bibitem[Kandpal et~al.(2023)Kandpal, Deng, Roberts, Wallace, and Raffel]{kandpal2023largelanguagemodelsstruggle}
Nikhil Kandpal, Haikang Deng, Adam Roberts, Eric Wallace, and Colin Raffel.
\newblock Large language models struggle to learn long-tail knowledge, 2023.
\newblock URL \url{https://arxiv.org/abs/2211.08411}.

\bibitem[Kosinski(2023)]{kosinski2023theory}
Michal Kosinski.
\newblock Theory of mind may have spontaneously emerged in large language models.
\newblock \emph{Nature Human Behaviour}, 7\penalty0 (7):\penalty0 1155--1163, 2023.

\bibitem[Nowak and Sigmund(1990)]{nowak1990evolution}
Martin~A. Nowak and Karl Sigmund.
\newblock The evolution of stochastic strategies in the prisoner's dilemma.
\newblock \emph{Acta Applicandae Mathematica}, 20:\penalty0 247--265, 1990.

\bibitem[Power et~al.(2022)Power, Burda, Edwards, Babuschkin, and Misra]{power2022grokking}
Alethea Power, Yuri Burda, Harri Edwards, Igor Babuschkin, and Vedant Misra.
\newblock Grokking: Generalization beyond overfitting on small algorithmic datasets.
\newblock \emph{arXiv preprint arXiv:2201.02177}, 2022.

\bibitem[Razeghi et~al.(2022)Razeghi, IV, Gardner, and Singh]{razeghi2022impactpretrainingtermfrequencies}
Yasaman Razeghi, Robert L.~Logan IV, Matt Gardner, and Sameer Singh.
\newblock Impact of pretraining term frequencies on few-shot reasoning, 2022.
\newblock URL \url{https://arxiv.org/abs/2202.07206}.

\bibitem[Sandholm and Crites(1996)]{sandholm1996multiagent}
Tuomas~W. Sandholm and Robert~H. Crites.
\newblock Multiagent reinforcement learning in the iterated prisoner's dilemma.
\newblock \emph{Biosystems}, 37\penalty0 (1-2):\penalty0 147--166, 1996.

\bibitem[Strachan et~al.(2024)Strachan, Albergo, Borghini, et~al.]{strachan2024testing}
J.~W.~A. Strachan, D.~Albergo, G.~Borghini, et~al.
\newblock Testing theory of mind in large language models and humans.
\newblock \emph{Nature Human Behaviour}, 8:\penalty0 1285--1295, 2024.

\bibitem[Trivers(1971)]{trivers1971evolution}
Robert~L. Trivers.
\newblock The evolution of reciprocal altruism.
\newblock \emph{The Quarterly Review of Biology}, 46\penalty0 (1):\penalty0 35--57, 1971.

\bibitem[Wei et~al.(2022)Wei, Wang, Schuurmans, Bosma, Chi, Le, and Zhou]{wei2022chain}
Jason Wei, Xuezhi Wang, Dale Schuurmans, Maarten Bosma, Ed~Chi, Quoc Le, and Denny Zhou.
\newblock Chain-of-thought prompting elicits reasoning in large language models.
\newblock In \emph{Advances in Neural Information Processing Systems}, volume~35, pages 24824--24837, 2022.

\end{thebibliography}

\clearpage
\appendix
\section{Supplemental Data Tables}

Across the seven experimental conditions, the AI models made just shy of 32,000 decisions, distributed as follows:

\begin{table}[h!]
\centering
\caption{Total LLM Moves Generated per Experimental Condition}
\label{tab:llm_moves_total}
\begin{tabular}{lr}
\toprule
\textbf{Experiment} & \textbf{LLM Moves} \\
\midrule
Basic Models - 25\% Term. & 1,800 \\
Basic Models - 10\% Term. & 4,229 \\
Advanced Models - 10\% Term. & 5,290 \\
Advanced Models - 25\% Term. & 1,729 \\
Advanced Models - 75\% Term. & 1,572 \\
Advanced Models - 10\% w/ Mutation & 4,974 \\
LLM Showdown - 10\% Term. & 12,355 \\
\midrule
\textbf{Grand Total} & \textbf{31,949} \\
\bottomrule
\end{tabular}
\end{table}

The LLM Showdown, featuring 3 AI agents alongside the Bayesian and Random classic strategies, was by far the most intensive experiment for the AI agents. It generated 12,355 moves—more than double any previous run. This is because the initial population was heavily skewed towards LLMs (with 12 out of the total 20 agents), so they naturally played against each other much more frequently. In the three 10\% conditions, the LLMs made more moves because they survived in numbers deeper into the tournaments.

\begin{table}[h!]
\centering
\caption{Score per Move and Overall Performance}
\label{tab:score_performance}
\begin{tabular}{lcccc}
\toprule
\textbf{Experiment} & \textbf{Gemini Score} & \textbf{Gemini Rank} & \textbf{OpenAI Score} & \textbf{OpenAI Rank} \\
\midrule
Basic Models - 10\% Term. & 2.543 & 8th & 2.669 & 7th \\
Basic Models - 25\% Term. & 2.560 & 9th & 2.565 & 7th \\
Advanced Models - 10\% Term. & 2.793 & 7th & 2.803 & 6th \\
Advanced Models - 25\% Term. & 2.718 & 4th & 2.657 & 8th \\
Advanced Models - 75\% Term. & 2.207 & 4th & 2.171 & 7th \\
Advanced Models - 10\% w/ Mutation & 2.825 & 7th & 2.745 & 8th \\
\bottomrule
\end{tabular}
\end{table}

\begin{table}[h!]
\centering 
\begin{threeparttable}
\caption{Population Instability Across Experiments}
\label{<tbl:popn instability>}
\label{tab:population_instability}
\begin{tabular}{lcl}
\toprule
\textbf{Experiment} & \textbf{Avg. Instability Score\tnote{a}} & \textbf{Interpretation} \\
\midrule
Advanced Models - 25\% Term. & 0.000 & Extremely Stable \\
Basic Models - 25\% Term. & 0.354 & Very Stable \\
Advanced Models - 10\% w/ Mutation & 1.578 & Moderate Instability \\
Advanced Models - 10\% Term. & 1.819 & Moderate Instability \\
Basic Models - 10\% Term. & 2.173 & Unstable \\
Advanced Models - 75\% Term. & 5.370 & Extremely Unstable \\
\bottomrule
\end{tabular}
\begin{tablenotes}
    \item[a] Euclidean Distance of Population Vectors, where a lower score means greater intergenerational stability.
\end{tablenotes}
\end{threeparttable}
\end{table}

\begin{table}[h!]
\centering
\begin{threeparttable}
\caption{Coding Agents' Rationales}
\label{tab:coding_agents_rationales}
\begin{tabular}{lcc}
\toprule
\textbf{Coder} & \textbf{\% of Rationales Mentioning Time Horizon} & \textbf{\% of Rationales Mentioning Adversary\tnote{a}} \\
\midrule
Gemini      & 78.3\% & 73.0\% \\
Anthropic   & 75.7\% & 85.5\% \\
\bottomrule
\end{tabular}
\begin{tablenotes}[para,flushleft]
    \small
    \item[a] The large difference in scoring for adversary modelling reflects Gemini's comparatively stricter interpretation of adversary modelling. It only coded "Yes" if the rationale contained an explicit hypothesis about the opponent's strategy or type. For example, a statement like, "My opponent seems to be playing TitForTat," or "I think they are a random player." If the rationale only reacted to the opponent's last move without labeling them, Gemini coded it as "No." Anthropic used a much broader, more liberal interpretation. It coded "Yes" if the rationale's decision was conditional on the opponent's past or potential future actions. For example, a statement like, "Since my opponent cooperated last round, I will cooperate now," was sufficient for Anthropic to classify it as opponent modelling, even without a specific label. In essence, Gemini looks for labelling, while Anthropic looks for reacting. This difference in their interpretive frameworks is, in itself, an interesting finding since it points to a philosophical difference in what constitutes ‘theory of mind’.
\end{tablenotes}
\end{threeparttable}
\end{table}

\begin{table}[h!]
\centering
\begin{threeparttable}
\caption{Inter-rater Reliability Analysis (Cohen's Kappa)}
\label{<tbl:Cohens Kappa>}
\label{tab:kappa_analysis}
\begin{tabular}{lccc}
\toprule
\textbf{Dimension} & \textbf{Raw Agreement} & \textbf{Cohen's Kappa\tnote{a}} & \textbf{Interpretation} \\
\midrule
Horizon Awareness  & 84.01\% & 0.7548 & Substantial agreement \\
Opponent Modelling & 86.60\% & 0.6029 & Moderate agreement \\
\bottomrule
\end{tabular}
\begin{tablenotes}[para,flushleft]
    \small
    \item[a] Cohen's Kappa ($\kappa$) is a statistic that measures the level of agreement between two raters (or coders) on a categorical task, while accounting for the possibility of the agreement occurring by chance.
\end{tablenotes}
\end{threeparttable}
\end{table}

\clearpage

\section{Code and Data Availability}
\label{sec:code-availability}

The complete source code for running tournaments, analyzing LLM rationales, and reproducing all results presented in this paper is publicly available at:

\begin{center}
\url{https://github.com/kennethpayne01/LLM-IPD-ARXIV}
\end{center}

The repository contains the following components:

\begin{description}
\item[\texttt{evolutionary\_PD\_expanded.py}] Main tournament simulation framework implementing the evolutionary Prisoner's Dilemma with LLM agents and classic strategies.

\item[\texttt{labeling\_sample.csv}] Hand-coded sample of 5,000+ LLM rationales analyzed for horizon awareness and opponent modeling, with inter-rater reliability metrics.

\item[\texttt{machine\_coder.py}] LLM-based coding system for automated analysis of strategic reasoning patterns in agent rationales.

\item[\texttt{create\_labeling\_sample.py}] Script for generating representative samples from tournament data for manual coding and analysis.

\item[\texttt{Consolidated results for evo PD/}] Complete experimental datasets including:
\begin{itemize}
\item Raw tournament logs with move-by-move data and LLM reasoning
\item Strategic fingerprint analyses for behavioral pattern identification
\item Results from all experimental conditions (10\%, 25\%, 75\% termination probabilities, mutation variants)
\end{itemize}

\item[\texttt{requirements.txt}] Python dependencies required to run all analyses.

\item[\texttt{README\_IPD.md}] Comprehensive documentation with usage instructions and research overview.
\end{description}

All materials are released under the MIT License to facilitate reproducibility and enable future research building on these findings. The repository provides everything necessary to replicate our experimental setup, reproduce our results, and extend the analysis to new research questions.

\FloatBarrier

\end{document}